# NeuroAI and Beyond

Jean-Marc Fellous[1], Gert Cauwenberghs[1,2], Cornelia Fermüller[3], Yulia Sandamisrkaya[4] and Terrence Sejnowski[1,5]

1. Institute for Neural Computation, University of California, San Diego, La Jolla, CA, USA.

2. Department of Bioengineering, University of California, San Diego, La Jolla. CA, USA.

3. Institute for Advanced Computer Studies, University of Maryland, College Park, MD, USA.

4. Institute of Computational Life Sciences, Zurich University of Applied Sciences in Wädenswil, Switzerland.

5. The Salk Institute for Biological Studies, La Jolla, CA, USA.

**Abstract**

Neuroscience and Artificial Intelligence (AI) have made significant progress in the past few years but have only been loosely inter-connected. Based on a workshop held in August 2025, we identify current and future areas of synergism between these two fields. We focus on the subareas of embodiment, language and communication, robotics, learning in humans and machines and Neuromorphic engineering to take stock of the progress made so far, and possible promising new future avenues. Overall, we advocate for the development of NeuroAI, a type of Neuroscience-informed Artificial Intelligence that, we argue, has the potential for significantly improving the scope and efficiency of AI algorithms while simultaneously changing the way we understand biological neural computations. We include personal statements from several leading researchers on their diverse views of NeuroAI. Two Strength-Weakness-Opportunities-Threat (SWOT) analyses by researchers and trainees are appended that describe the benefits and risks offered by NeuroAI.

# Contents

# Introduction

Learning is central to biological systems, it is at the core of machine learning algorithms, and data-driven deep learning fuels a large number of AI approaches. Learning occurs over a wide range of biological time scales, from seconds for sensory habituation, hours for motor adaptation and long-term working memory, years for lifelong learning, and millennia for evolution. In large language models (LLMs), there are only two time scales: A long and expensive pretraining, and inference, which is fast but does not update the weights in the network.

Although NeuroAI focuses on brains, bodies are crucial for grounding interactions in the world. Nature has evolved many types of bodies for different environments in the air, water, and land. Brain architectures co-evolved with bodies; neuroethological studies will help us adapt AI to diverse environments. We aim to derive general principles from this diversity that can help us design brain architectures for NeuroAI that are just as diverse.

*Will giving AI a body that interacts with the world be a path toward Artificial General Intelligence (AGI)?* Humans learn by interacting with the world. Although large language models are pretrained based only on text, they seem to have an internal model of the world and humans. However, not everyone agrees, and brain-body interactions might provide a missing link to AGI.

*Are there fundamental limits on language AI, and should Large Language Models have a developmental phase similar to early childhood?* Humans have a long development during which they learn about the world and are aligned to their culture through reinforcement learning and imitation. Large language models are only fine-tuned after pretraining, which is inefficient and makes them inflexible.

*How can LLMs achieve lifelong, continual learning by updating long-term memory while preserving old memories?* LLMs do not have long-term memories, and attempts to introduce continual learning lead to degradation of the initial training. In contrast, human memories are updated and consolidated daily during sleep, minimally interfering with older memories. AI assistants and tutors will need a long-term memory that can be updated.

*Do robots need a layered, distributed, adaptive, and robust operating system similar to humans?* The brain's motor system has many layers of control, with time scales of milliseconds in the spinal cord to long-term planning in the cortex, coordinated by an operating system made of multiple interacting brain areas. Robots could be more flexible and robust by adopting layered, distributed control.

*Can we shift from energy-hungry digital simulations of large AI systems to ultra-low power neuromorphic hardware on edge devices like glasses and smartphones?* One of the reasons why LLMs are power-intensive is because they are simulated on digital computers. In nature, learning and computation are performed at ultra-low power by embedding computation in a physical substrate. Brains take computing down to the molecular level. This is possible in silicon with analog VLSI processing, a technology that has matured since it was introduced by Carver Mead in 1986.

The science and engineering of learning is still in its infancy, and the bridge between AI and brains is accelerating advances. Just as the industrial revolution greatly enhanced our physical power, the NeuroAI revolution could significantly enhance our cognitive power and give us a better understanding of ourselves. Our trainees will inherit powerful tools and a new conceptual framework for this transformation.

# 1. Open Questions in Embodied Cognition and Computation

## What is embodiment, in the context of NeuroAI?

An *embodied system* is a physical, biological, or simulated system that is autonomous and situated in feedback with an environment, in that it receives inputs from the environment through sensors and acts upon its surroundings through actuators. In general, this feedback implies that an embodied agent has a body (a morphology with physical constraints) that is capable of interacting with the physics of its world as well as with other agents. This interaction necessarily involves perceptual and motor delays at multiple time scales, and effective embodied systems are designed to handle them using predictive algorithms. Crucially, embodied systems have an identity that persists through time. To be efficient and flexible, the morphology and the controller are co-designed or jointly evolved. As a consequence of feedback interactions, an embodied system may develop internal models of itself and of its environment, allowing it to "play forward in time" and plan through possible behavioral scenarios.

## Why is embodiment important for NeuroAI?

Allowing an agent to explore autonomously, interact, as well as modify and be modified by a complex environment will provide it with the experience necessary to construct an interactive understanding of that environment. Such an understanding will provide it with appropriate meaningful rewards and punishments that will allow it to generate flexible and adaptive behavior even as the environment (or other agents) pose challenges to it. Therefore, developing neural models (including those based on foundation models and large language models, especially if operating in a continual learning mode) within an embodied context could provide them with a natural path to capture the dynamics of the environment, generate rapid and effective responses, store past experiences, and plan future actions, at every timescale relevant for the agent.

## How are we going to do it?

New approaches to embodied cognition should embrace the use of (1) computational simulations to create biomechanically realistic embodied agents and physically realistic artificial worlds, (2) theoretical as well as "simplified" models capable of cognitive behavior, (3) experimental work exploring autonomy and feedback in a range of animals from experimentally-tractable invertebrate model systems to complex vertebrate systems, and (4) studies that investigate the role of embodiment in human language and cognition. Embodied systems should implement real-time solutions that focus on *satisficing* (in Herbert Simon's and bounded rationality sense) algorithms and on being good enough to function effectively in a changing environment, as opposed to taking a long time to calculate optimal solutions. In creating embodied agents, it is important to incorporate multiple timescales of interactions. These include short-term and effective responses to environmental challenges (e.g., rapid mechanical responses of the body to perturbations, reflexes to cope with irregular terrain during locomotion, defensive withdrawal responses to avoid damage, reactive repair/healing mechanisms), medium-term responses that may require comparisons of actual feedback to a desired motor output as well as the ability to make appropriate adjustments (e.g., corollary discharge), and then, because there are always delays from perception to action, the ability to make appropriate predictions (e.g., feedforward activation of circuits that are about to be employed, longer-term predictions of the consequences of different actions). Incorporating neuromodulation (such as produced by dopamine, serotonin or norepinephrine, to name only a few), which allows for the coordinated activation and modulation of different parts of the neural circuitry either in response to or in anticipation of environmental changes, will be crucial for effective coordinated behavior. Incorporating mechanisms for sensing painful stimuli or pleasurable sensations could serve as the basis for creating artificial emotions, which could significantly improve, guide and support currently purely cognitive algorithms.

## When is a good time to do it?

We believe that the right time to do this is now. Advances in technology are creating remarkable opportunities for embodied cognition. We have massive datasets including those from optogenetics, large-scale multi-site electrophysiological recordings, connectomics, and comprehensive behavioral tracking. How can we integrate and make sense of this data? **First**, we need appropriate *theoretical frameworks* for trying to understand the data, which will need to include the development of *general theories of embodiment*. Investigators have often focused on a "lexicon" for interpreting symbols, but, in embodied systems, developing a "praxicon" to represent movements and associated sensory inputs in an abstract sensorimotor space may be just as important. **Second**, we should integrate diverse data into embodied models that constitute "dynamic databases" that support tunable parameters corresponding to the experiments that could be done on the system and, thus, can be used for the design and prediction of outcomes of experimental manipulations. **Third**, we should develop these models such that they could have significant diagnostic and predictive power. Moreover, by running ensembles of such models with varying initial conditions, one could quantify uncertainty with respect to future behavior and further begin making short-term predictions of behavior from a neural state.

## What conditions will foster the development of embodied cognition and computation?

For success in such trans-disciplinary efforts, it will be important to train a new cohort of investigators that have expertise in biology, physiology, physics, cognitive science, engineering and computer science (among other disciplines). This calls for creating new interdisciplinary training programs that should engage several investigators from different disciplines, as well as further provide incentives for exchanges among laboratories of postdoctoral fellows and graduate students, exposing them for at least two to three months to different approaches, theoretical perspectives, and methodologies (e.g., over a summer, or a semester) in pursuit of tools, skills and conceptual frameworks that will strengthen the trans-disciplinary research effort of the involved laboratories. Furthermore, the scale of many of these efforts go well beyond the scope of a single laboratory and may therefore require larger-scale efforts that bring together a very large number of investigators, including both scientists and engineers. These problems have been dealt with by the particle physicists (e.g., CERN) and the astronomers (e.g., the Webb telescope), and their approaches to the problem (including historical lessons learned) may be a useful model to consider. Finally, it is important to take a very broad range of approaches, including not just bio-inspired approaches, but perhaps using biological materials and more abstract concepts derived from biology for creating the novel devices that will allow us to make progress. It will be also important to develop ethical and legal standards and best practices, as the technology and science of embodied cognition and computation move forward.

## Is embodiment a path towards general intelligence?

Humans and animals constitute proofs by existence that embodiment may play a crucial role in giving rise to general intelligence. Indeed, the only agreed-upon examples we have of intelligent agents are those that are embodied. If we describe intelligence more broadly as adaptive behavior that allows an organism to learn, survive and reproduce in a complex, changing environment, rather than circumscribing it purely to linguistic behavior, then it is clear that many animals show embodied intelligence. In addition, there is a clear evolutionary continuity between the kinds of complex behaviors shown by animals such as the octopus, dolphins, or chimpanzees with the more advanced behaviors observed in humans.

Taking all of these reasons together, we hypothesize that embodiment is necessary (but not necessarily sufficient) to create generally intelligent agents. Crucial features for intelligent embodied agents will be short-term adaptation, learning and memory, and the ability to reconfigure controllers in a multi-scale way over time. Embodiment may be key to move from *simulated* general intelligence (as LLMs may already be showing) to *actual* general intelligence.

## 2.   Open Questions in Language and Communication

Large language models (LLMs)—neural networks trained on massive amounts of textual data with a next or missing word prediction objective—have achieved human or super-human performance on diverse language tasks. Moreover, recent LLMs trained on non-language data (e.g., databases of computer code) and supplemented with reinforcement learning from human feedback (RLHF) achieve non-trivial performance on some forms of reasoning. However, even state-of-the-art LLMs and LRMs (large reasoning models) remain cognitively limited and differ significantly from human brains in the implementation of their abilities. In particular, current AI models are not biologically, cognitively, or developmentally plausible, and remain fragile in their performance on many cognitive tasks, affected by minor perturbations to which humans are robust.

If LLMs and LRMs and other AI models are to serve humanity in i) advancing our scientific understanding of human minds and brains, and of intelligence more broadly, and ii) improving human lives, we need to continue improving these models, including by using insights about the efficiency of biological brains. These two goals come with their own desiderata, constraints, considerations, and possible plans of action, so we discuss these two goals in turn.

Goal 1: To use AI models as models of human language and cognition, and to ask broad questions about the nature of intelligence

### How can we build more cognitively plausible models?

What degree of modularity is needed between language and reasoning systems? How do the differences between the (relatively) conscious declarative hippocampal learning system and the non-conscious procedural, basal ganglia learning systems inform our understanding of true chain-of-thought reasoning and consciousness itself? Should AI preserve modular architecture separating emergent reasoning from language  (Fedorenko and Varley, 2016; Fedorenko et al., 2024)?

### How can we build more developmentally plausible models?

The human brain coordinates specialized systems—language, memory, reasoning, social cognition—through robust interfaces. What can we learn from how infants acquire language efficiently from limited multimodal data, leveraging embodiment and social interaction, while large AI systems require vast text corpora? Building on philosophical traditions from Kant through cognitive psychology, should AI begin with innate mental structures and foundational concepts (space, time, quantity, relations) as scaffolding? As Cantlon & Piantadosi (2024) argue, uniquely human intelligence arose from expanded information capacity (Cantlon and Piantadosi, 2024) —how can this insight inform AI development? What intermediate representations beyond natural language are needed for unambiguous execution in sensorimotor space? Current approaches like reinforcement learning with world models represent first steps but may fall short, as human understanding encompasses tacit knowledge of actions, embodied perceptions, and affective experiences. How can AI develop richer representational frameworks integrating language with sensory, motor, and emotional dimensions?

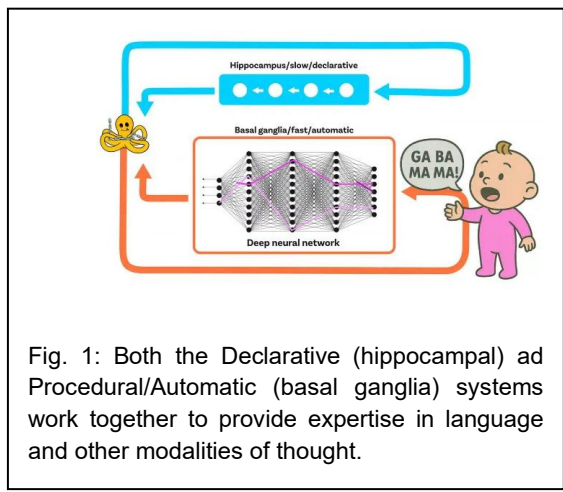

Fig. 1: Both the Declarative (hippocampal) ad Procedural/Automatic (basal ganglia) systems work together to provide expertise in language and other modalities of thought.

### What are the fundamental limits of language-based AI?

Language serves as a medium for expressing thought, both introspectively and communicatively, but is not synonymous with reasoning and thought. (Figure 1) While human language is inherently grounded in perception and action, with children learning it within social contexts, most LLMs are trained exclusively on text lacking this grounding—making them fragile when tested systematically. Do current AI systems' fluent outputs constitute genuine thought and creativity, or merely sophisticated patterns of recombination? How can new tokenizing approaches for sensorimotor and other modal signals enable training beyond text? Should AI integrate multimodal inputs for flexible, creative intelligence—potentially explaining the overlapped activity seen when human knowledge and reasoning engage cognitive and motor faculties? An often-neglected challenge is that cognition did not evolve in isolation but in constant interplay with other humans

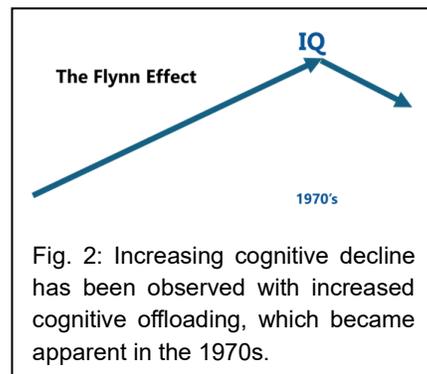

Fig. 2: Increasing cognitive decline has been observed with increased cognitive offloading, which became apparent in the 1970s.

Goal 2: To use AI models to improve human lives across diverse areas

## How can we prevent cognitive atrophy from AI reliance?

Declarative and procedural memory (Figure 1) must be exercised through retrieval and practice—how do we prevent constant offloading to external tools from disrupting schema formation and weakening flexible reasoning? (Oakley et al., 2026).The anti-Flynn effect (the decline in IQ scores observed in Western countries since the 1970s) appears to be correlated with increased cognitive offloading that began with the widespread use of calculators in the 1970s. (Figure 2)

## Can NeuroAI be used as a massive springboard for K-12 teacher training?

Could partnerships with initiatives like the National Academy for AI Instruction (Microsoft, OpenAI, and AFT) revolutionize education?

## How can we apply neuroscience insights about cognitive biases and motivated reasoning to develop better alignment strategies?

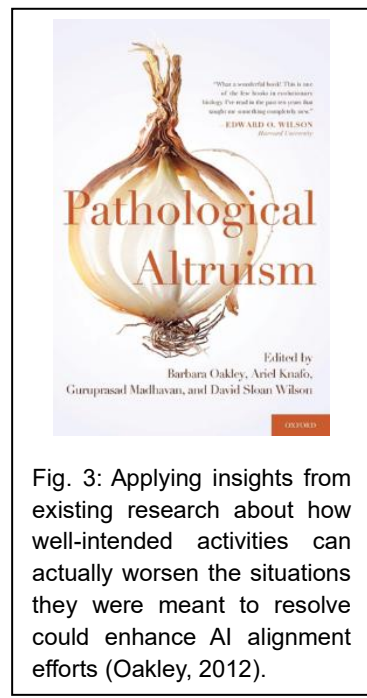

Fig. 3: Applying insights from existing research about how well-intended activities can actually worsen the situations they were meant to resolve could enhance AI alignment efforts (Oakley, 2012).

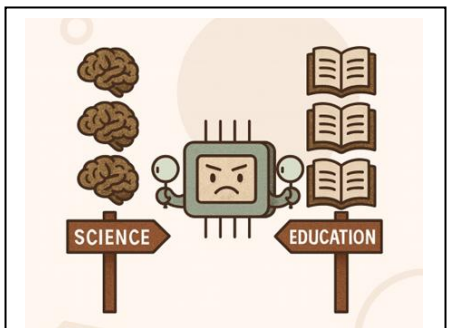

Fig. 4: Large scale groups like those researching Alzheimer's disease, or schools of education, can sometimes lock into a paradigm (e.g. amyloid plaques are the cause of Alzheimer's, whole language is the best way to teach children to read).

Understanding how human brains can exhibit pathological altruism (well-intended actions that cause harm) can inform AI safety approaches that avoid similar failure modes — for example, preventing overcorrection in content moderation that leads to absurd outputs (Figure 3).

## How can we leverage NeuroAI's interdisciplinary nature to identify "paradigm cartels"?

Just as neuroscience revealed how entrenched theories (like the amyloid hypothesis) held back Alzheimer's research for decades, how can NeuroAI's cross-domain perspective help identify where siloed thinking limits progress across many fields? (Figure 4)

# 3.   Open Questions in Robotics

## Efficiency: While Deep Learning (DL) is extremely powerful at learning statistics from data, what is the cost? Can we take inspiration from the brain on how to do so more efficiently?

Efficiency is in being able to accomplish a task well enough and with low energy expenditure. Part of this relies on the ability to learn over time, with few samples and adapt to distribution contingencies.

Achieving efficiency in robotics requires taking a holistic approach, that encompasses the co-design of the computation and the sensors and actuators. Co-localizing memory and computation, beyond the von Neumann computing paradigm cuts the cost of moving data around. Moving beyond the feedforward structure of DL by integrating the massive feedback towards sensory areas observed in the brain will enable the implementation of a predictive machine that does not require the expensive speed up of computation by powerful and fast computing devices. Understanding how the brain computes with slow and unreliable computing elements can lower the cost of developing extremely fast and precise hardware. Using the multifaceted concept of embodiment by exploiting the hardware and the (purposeful and mechanic) interaction between the body and the environment will offload part of the computation to the materials, sensors and actions.

## Safety & Reliability: How can we design robots that are guaranteed to be (certifiably) safe?

Safety is both in the decision-making, to ensure that behavior is correct, and that the movements of the robot are physically safe (non-physically damaging). Their behavior must be robust and reliable in all conditions, capable of managing out-of distribution/non-standard occurrences.

Robotic safety also encompasses verbal safety, i.e. non-toxic verbal interactions. Control theory can be applied to LLMs and DL in general for controlling toxicity in text. In robots, hierarchical DL architectures can give flexibility but cannot guarantee safety in the same way as the Model Predictive Control can; there is a need to define a trade-off between the two in each layer.

## Biological grounding: How can we better exploit lessons from biology to develop efficient, safe, reliable robots?

Mainstream robots, even animaloids and humanoids that share a physical resemblance to their biological counterpart, are built around hardware and software infrastructure that is far from biology. While we should not aim at faithfully replicating biology, more biologically-inspired computational and design principles could be introduced to fully exploit the brain-body synergy that characterizes biological systems. To achieve this aim, the current robot architectures might need to be challenged. To do so, there is the need for a holistic, unified framework for robotics that integrates hardware and software from the ground up, i.e., from low-level materials and haptics to high-level cognitive functions, moving away from the current rigid and segregated approach. Current approaches often assemble pre-existing, isolated modules (e.g., Simultaneous Localization and Mapping (SLAM), perception, motion planning, inverse kinematics, actuation) that were researched and developed over decades often in isolation without a globally optimized architecture. The modularity of embodied systems is often an illusion. In biological systems and robust robots, replacing a key component like a camera or an "eye" is not trivial, indicating deep integration rather than true modularity. The question is 'what would a global "operating system" that supports this holistic concept for robotics look like?'.

## Ideas and directions can be classified into main principles

<u>Principle 1: Hierarchical Stack</u>

Hierarchical stack with layers (and levels) and a principled framework to design each layer are needed not only for the "brain", but also for the body, starting from the materials. Whatever architecture is devised must be hierarchical, with (slower) deliberative high-level control of goals that coordinate faster intermediate levels that ground those goals into actions that depend on the context and low levels that must be fast for actuation and adaptation. In the brain, the layers of the hierarchy are physically instantiated in different and specialized brain areas and are strongly tied to the actuation elements, with a lot of local feedback to the level of sensing. We need controllers at very different temporal scales for adaptation and flexibility. For example, the vestibulo-ocular reflex (VOR) is continuously recalibrated (e.g., when wearing glasses), the cerebellum detects a blur and changes the controller's gain slowly, but then it is learnt by the vestibular nucleus and it becomes extremely fast to change from one setting to the other from context detected by high level brain areas. Controllers should therefore adapt very slowly to new conditions but be able to quickly switch when recognizing an experienced condition. This hierarchical structure for planning and control remains an unsolved problem in robotics. While robots use hierarchical planning, it is typically fixed and designed by hand. The goal is to incorporate the learning and flexibility demonstrated by DL. While DL is flexible, it is also not certifiable in terms of guaranteed safety, while less flexible robotic approaches are intrinsically safe. The robotic hierarchy envisioned will probably rely on a hybrid approach where lower levels rely on the latter and higher levels rely on DL.

<u>Principle 2: Distributed Control</u>

Mainstream robots are rigid and over-actuated and use a monolithic control architecture to simplify their control and guarantee safety, at the cost of energy, flexibility and adaptability. One pathway is to look for artificial muscles and linear motors, replacing rotational motors, using springs and dampers that are **seamlessly controlled** by the 'brain' of the robot. Controllers should be adaptable, but at a very long time scale. Biological motor systems feature sensorimotor control loops and feedback at every level of the hierarchy. Current robotics architectures often fail to account for this kind of distributed control. Recent advances in control theory can now mathematically handle the sparse signals and time delays inherent in distributed control, but these methods need to be more broadly applied and available to the research community as off-the-shelf tools to support the design of new robotic control stacks. Distributed controllers at the local level can potentially obviate the need for complex calculations like inverse kinematics, with higher-level centralized systems acting as coordinators.

<u>Principle 3: Full-Stack Codesign</u>

Robotics requires a full-stack approach where the robot is defined from its base materials all the way to its highest-level computation. **Co-design of Body and Controller:** The physical design of the robot's body is critical and should be developed in tandem with its control system to make control easier. This includes choices in materials (e.g., soft robotics for safety), actuators (e.g., linear actuators or artificial muscles over rotational motors), and the use of passive dynamics (springs, dampers). **Feedback Integration:** The entire stack must be connected by feedback loops. While current DL focuses on improving prediction, robust systems must incorporate information from sources like haptics and proprioception, as well as feedback from the brain to the sensory system through contextual tuning of receptors and signal filtering and actions. We therefore need DL that can integrate massive feedback.

<u>The Need for Standards</u>

To enable community-wide progress, a principled, layered architecture requires standards for device and software communication, analogous to standard operating systems and hardware interfaces in computing (e.g., x86/ARM, Linux, USB). An "OS-equivalent" for the mechanical, computational, and fabrication aspects of robotics could foster the kind of diverse innovation seen in biology, where a single piece of machinery (the ribosome) produces enormous protein diversity. The final goal should be an end-to-end supporting infrastructure.

## From NeuroAI to robotics: What should robotics learn from neuroscience?

Neuroscience and NeuroAI can offer a theory-grounded architecture to build effective robots. Neuroscience can offer Architectural Blueprints (to the benefit of robotics and NeuroAI)

**Hierarchical and Distributed Control:** Perception utilizes hierarchical representations (as DL has shown for the visual dorsal pathway), and a similar approach should be applied to motor control. The physical separation of these layers in the brain (e.g., cortex, cerebellum, spinal cord) suggests a robust design pattern.

**Specialized Pathways:** The brain uses different pathways for different tasks. For example, the visual system has a fast pathway for posture and movement control and a slower one for object recognition. This suggests that robotic systems could benefit from specialized, parallel processing streams rather than a single monolithic one.

**Learning and Adaptation:** The brain excels at learning "intuitive physics" over which it builds a world model. New DL models aim at learning to predict future states from past states and actions. This learning happens in stages, similar to infants: first through passive observation to build a good world representation, and then by incorporating actions to learn cause and effect. This is a path to move beyond Large Language/Vision Models: LLMs and Vision Language and Actions models (VLAs) are not a silver bullet for robotics. They operate in a discrete, turn-based space, while robotics requires real-time control in a continuous world. Furthermore, VLA architectures are often not "physical"—they are expensive to train, tied to specific hardware, and lack the safety, predictability, and reliability required for physical interaction.

**Plasticity and Continuous Calibration:** Biological systems exhibit plasticity at all levels. The VOR is a prime example. The cerebellum constantly detects errors (retinal slip) and slowly recalibrates the gain of the reflex, which is then learned by the vestibular nucleus for extremely fast execution. This model of slow adaptation combined with fast execution is a powerful lesson for robotics.

**Feedback, Not Just Prediction:** The brain is a closed-loop system with massive feedback, from efferent copies of motor commands to feedback from high-level cortical areas to early sensory areas (e.g., LGN). Transformers use attention as a form of feedback, but it's not the rich, multi-layered feedback seen in the brain. Incorporating true, delayed, and sparse feedback signals is critical.

**Co-design of Brain and Body:** Biology simplifies control through clever body design. This includes using tendon-driven limbs, passive dynamics (springs and dampers), and soft materials. This is in contrast to many modern robots, which are stiff, over-actuated, and energy-inefficient to make them mathematically simpler to control. Humans are highly dexterous due to massive sensing in the periphery (e.g., skin). Haptics, which integrates touch, vision, and proprioception, is a crucial area where robotics can learn from neuroscience. Developing robust, sensorized skins with embedded neuromorphic processors is a key research direction. The brain doesn't process raw pixel values. The retina performs significant pre-processing and encodes information efficiently. Neuromorphic sensors like Dynamic Vision Sensors (DVS) are inspired by this, encoding changes and events rather than static frames, which is a more efficient input for many robotic tasks.

**Integrating Control Theory and AI:** A promising path forward is to merge the strengths of classical control theory (which provides safety and predictability) with modern AI. New mathematical tools in control theory can handle complex constraints and represent high-level intelligence within the state space of a dynamical system, providing a universal framework to integrate learned models with low-level controllers.

## From robotics to NeuroAI: Is the robotics field relevant to the development of the next generation of NeuroAI systems?

Robots can be used to uncover challenges that neural systems face, guiding the development of NeuroAI. Building physical robots can provide unique insights, testbeds, and constraints for neuroscientific theories and artificial intelligence systems. Real-world constraints such as wiring limitations (e.g., the size of an optical nerve), energy consumption, heat dissipation, delays in computation, communication and actuation, and physical safety are often abstracted away in simulations but become paramount in robotics, providing concrete problems that can inform models of computation. The act of building a robot forces a precise definition of concepts. While robotics is a subset of embodiment, its specific requirements can help delineate what aspects of embodiment are essential for different cognitive functions. Building a platform raises critical questions about applying biological models directly. For example, if a robot's embodiment is fundamentally different from an animal's, using a neural architecture that does not match the body's constraints may be inefficient or ineffective, pushing neuroscientists to consider the tight coupling between brain and body. Robots serve as platforms to test neuroscientific hypotheses about control, perception, and embodiment. By building a system based on a biological principle, researchers can validate its sufficiency and explore its limitations. Robotics can explore a wider design space than biology allows. By building robots with different sensors (e.g., wider Field of View), actuators, and body plans, we can study how these physical differences impact control strategies and intelligence, offering a comparative perspective to neuroscience.

Computationally, NeuroAI models that will run on robots will have to handle sensory signals that are partial, redundant, noisy, ambiguous, changing and multimodal. Sensing, decision-making and actions are continuous processes and local and on-demand learning needs to rely on small data, sparse supervision or reward. NeuroAI for robotics will benefit from architectural diversity, relying on different ways to encode and represent information (not just vectors/tensors) and on different "microcircuits" for different functions, all in a unified architecture.

## 4. Open Questions in Learning in Humans and Machines

### Interaction and Temporal Dynamics: *How can we implement brain-inspired mechanisms to enable AI to learn throughout its lifetime from ongoing real-world interactions?*

To enable AI to learn continuously throughout its lifetime from real-world interactions, we can implement brain-inspired mechanisms like neuromodulation and varied synaptic timescales. Neuromodulation can act as a gating mechanism, signaling to the system when and what to learn based on context, novelty, or internal states, thereby balancing plasticity with stability. Paired with this, incorporating a spectrum of synaptic timescales—some connections that change quickly for rapid, short-term adaptation and others that change slowly for stable, long-term knowledge—would equip the AI to effectively process continuous streams of integrated multimodal data. This dual architecture would allow the system to capture the multiple levels of temporal correlation inherent in real-world information, creating a more robust and flexible lifelong learning agent.

### Architecture Design for Robustness: How can AI architectures be enhanced in order to improve the robustness of systems in the face of perturbations?

To improve the robustness of AI systems in the face of perturbations, we can enhance their architectures by integrating specialized components that are functionally analogous to the brain's circuits. For instance, incorporating a module for predictive control, much like the cerebellum, would allow an AI to anticipate and smoothly correct for real-time disturbances. This could be combined with a system for rapid, high-capacity associative memory, inspired by the hippocampus, enabling the system to quickly recall relevant past experiences to inform its response to novel or unexpected situations. It's crucial that this approach is driven by concrete functional hypotheses about why these brain structures confer resilience, rather than simply copying their architecture for its own sake. By thoughtfully combining these specialized functions, we can build more resilient and adaptive AI systems that can better handle the unpredictability of the real world.

## Modular Learning: Can we develop AI systems that are better at respecting constraints that exist on multiple levels of the world model?

We may be able to develop AI systems that better respect multi-level constraints by creating hierarchical modular learning structures akin to neocortical microcircuits. Each module in this hierarchy could optimize local losses, handling specific, low-level constraints, while also flexibly incorporating global losses that represent the system's overarching objectives. To make this work, it's essential to also model how cortical and subcortical loops coordinate and control these various modules during learning. These loops would act as a control mechanism, ensuring that the local learning processes align with the global constraints of the world model. This approach would enable the AI to build a more coherent and robust internal representation of the world that simultaneously satisfies rules at multiple scales.

## Development and Evolution: How can we avoid the need to design and retrain massive foundation models from scratch?

To avoid the need to design and retrain massive foundation models from scratch, we can draw inspiration from biological development and evolution. Instead of engineering a fixed, final architecture, we could leverage principles from development to enable architectures to self-organize based on a sequential program specified by a compact set of rules, much like a genome. The primary task is then to find pathways toward non-random weight initializations that are inherently useful for AI. This would likely require a process that combines something akin to evolution, which would search for effective rule sets, with a developmental process that grows a structured network from those rules. This approach would replace the current paradigm of random initialization with a structured starting point that already encodes valuable priors, potentially bypassing the need for exhaustive retraining.

## Prediction and Understanding: How can we build agents with common sense?

To build AI agents with common sense, we should shift our focus from passive prediction on static data to active learning through action-conditioned prediction. This approach requires an agent to actively build its world model by observing the consequences of its own behavior, rather than simply processing a fixed dataset. However, a significant challenge in this pursuit is that it is not yet clear how to properly assess whether an agent, be it artificial or natural, has actually learned a robust world model. Progress, therefore, depends not only on designing these active learning systems but also on concurrently developing the principled analytical tools needed to quantify and validate the common-sense knowledge they acquire.

## Energy Efficiency: How can we improve energy efficiency during training?

To significantly improve energy efficiency during AI training, we must revolutionize the underlying hardware by adopting the brain's core principle of co-locating memory and computation. By using techniques like 3D chip design to bring processing and memory closer together and by designing for sparsity, we can drastically cut the energy wasted shuttling data back and forth, which is a major bottleneck in current architectures. The ultimate goal is to create systems where the physics of the chips lead naturally to the optimization of the system, meaning the hardware's own physical dynamics contribute directly to the computational task. This moves beyond simply running algorithms on a chip to designing a chip whose physical properties inherently solve the problem in a more efficient way.

# 5. Open Questions in Neuromorphic AI Engineering

● We live in a time of unparalleled computing and data advancements in spheres of machine learning and the understanding of brain structure and function. Never before has humanity been able to record very high-resolution neural activity as we are doing today. Furthermore, the symbolic and generative modeling abilities of today's AI models are impressive.

● AI models have benefited from the fundamental understanding of neural and cognitive function, and we are at a juncture where the converse can become equally true. Modern AI language models have demonstrated the ability to reason and synthesize long-range contextual meaning from high dimensional text, speech and visual input. Cerebral data are inherently high dimensional regardless of the level of abstraction – from biomolecular pathways, neural dynamics to population/ region activities. Future LLMs can potentially distill novel insights into neurophysiology, finding the needles buried in the haystack of neural signals.

● As a field that has evolved in close tandem with neuroscience, the principles of neuromorphic engineering will continue to influence the design of efficient hardware and algorithms to meet the sustainability concerns of today's AI models. We need to think about the fundamental biological plausibility of current AI models and opportunities to move these models closer to the biological brain, keeping clearly in view the differences in the biological and electronic device technologies and industrial requirements. The novel silicon (and other emerging materials) and photonic technologies for improving computation and communication bottlenecks of current hardware are quickly advancing. Monolithic 3D integration of colocalized compute and memory are also of great importance. The co-design of hardware and algorithms that enforce various kinds of sparsity (be it spatial, temporal, quantization, pruning, etc) and leverage the efficient shuttling of data for such non-contiguous data organization that such sparsity bring will likely improve existing approaches. For context, GPUs are excellent for dense linear algebra but are inefficient for sparse equivalents. Thus, there is an opportunity for neuromorphic approaches to offer sparsity-aware solutions. Sparse computation in this 3D compute-paradigm could potentially obviate the need to handle volumetric heat dissipation that would ordinarily accompany dense 3D computations.

● There are some concerns on improving accessibility to DRAM interfaces. to allow future neuromorphic hardware to transcend the limitations of single-chip SRAM memory capacity and to allow scaling of model translation efforts to current high-parameter models. Such models will include the obvious large language models as well as large-scale emulations of brain connectomes such as the recent fly connectome emulation in the Intel Loihi 2 chip (achieved by tediously networking several low-memory neural processing cores instead of a single system-on-chip solution equipped with high memory capacity to house the irregular graphic connectome structure). As dire efforts continue to keep Moore's law alive for immediate impact in the efficient design of AI hardware accelerators, older technology nodes are still useful for implantable medical device ASICs as they often require low-noise analog sensing and overall reliability to meet strict regulatory approval. Heterogeneous integration methods that combine advanced nodes for digital compute and older standard nodes for sensing are equally viable at least in the short-term.

● In general, we advocate for an approach to neuromorphic AI engineering with co-design of Neuroscience, Algorithms, Hardware and Sensors (Figure 5). We also advocate for support for educational efforts, such as the 30+ year old NSF-sponsored Telluride Neuromorphic Cognition Workshop, to continue catalyzing such intersectional research within the short to long-term horizons and to counter the massive People Republic of China investments in AI, robotics, and neuroscience in the last year.

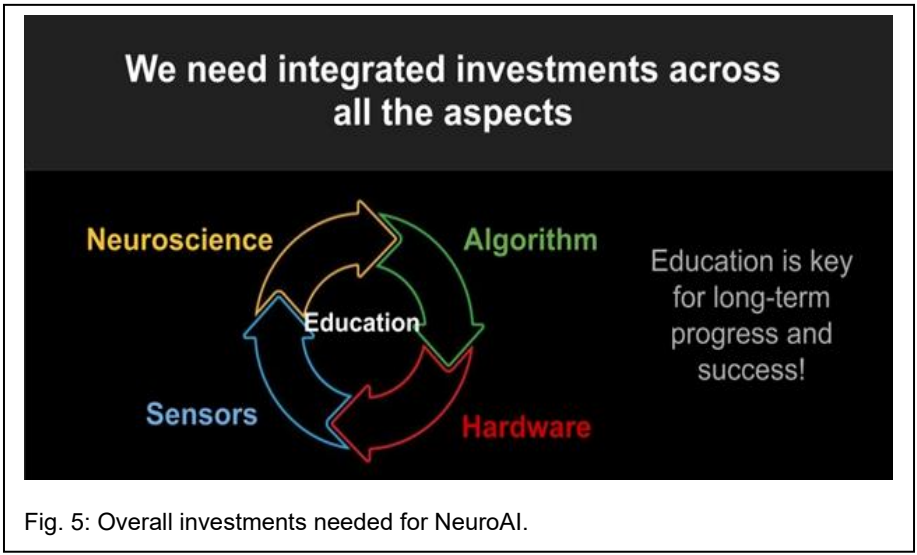

Fig. 5: Overall investments needed for NeuroAI.

○ *Neuroscience:* A neuromorphic solution is a 'brainchild' of neuroscience, as such the level of biological plausibility exhibited by neural emulations will only be as good as the current known neuroscience. On the other hand, advancements in cognitive understanding will emerge from the use of such derived engineered systems – essentially a chicken and egg problem of the substrate building the tool and vice versa. Advancements in both domains will come from iteratively updating each as new insights emerge in the neuroscience-neuromorphic loop. Call it analysis by synthesis and synthesis from analysis.

○ *Algorithms:* Bio-inspired learning algorithms share a similar fate, with new information about the brain, more and more efficient learning schemes will emerge for adoption in AI foundational models. In corollary, these models will be effective tools for choosing optimal hardware and sensors that will potentially reveal new cognitive insights as they advance artificial recognition in the wild.

○ Hardware: Intelligence can be abstracted to various material substrates and systems but optimality of a hardware solution for a chosen task will require a clear understanding of the algorithm being considered. E.g. probabilistic learning models (algorithm) emulating synaptic stochasticity at scale (neuroscience) can leverage transistor process-voltage-temperature (PVT) variations.

○ Sensors: Constraining redundancies at the earliest stage of information flow will dictate downstream efficiency. A tight colocalization of sensing and processing will address issues of energy-efficiency and latency in end-to-end always-on learning and inference engines.

## How to best draw inspiration from natural intelligence?

<u>How Neuro is current AI?</u>

There are several neurally-inspired elements and operations of the current AI model paradigm such as nonlinear activations (ReLU, softmax, …) and convolutional operations like receptive fields, max and average pooling (both inspired by complex cells in the hierarchy of processing in the visual cortex). Other promising concepts such as sparsity and attention mechanisms expressed in current AI models have strong bases in various cortical pathways. It is unsurprising that the 'secret sauce' to transformer architecture today is attention. This version of attention while being rather abstract is in essence akin to cortical processes such as contextual modulating (e.g. von der Heydt work) where local circuits influence novel visual stimuli arriving in the cortex. There are other biologically inspired learning frameworks such as reward/ reinforcement learning with obvious semblances to reward processing by the basal ganglia and dopaminergic systems. Other concepts like probabilistic computation have parallels in cerebral processing considering the prevalence of stochasticity in the brain especially at the synapse level. Other emerging foundational models such as recurrent neural networks and state-space models are loosely inspired by the neural dynamics of leaky integrator neurons.

<u>How *not* like Neuro is current AI?</u>

There are however deviations from biology in current models. First, the tokenized approach to cognition is very different from the way information is processed in the brain. Natural and neural processes are rather continuous in space-time and gracefully adopt sparsity where essential to meet the tight energy budget. On the other hand, LLMs are riddled with a plethora of dense matrix-matrix multiplications particularly in the cascade of multi-head attention units. Second, the incorporation of feedback mechanisms in current AI is limited. Most models while being plausibly hierarchical are purely feedforward and stateless. Artificial neural network (ANN) models often compensate for this lack of statefulness by achieving depth spatially rather than temporally as done by the cortex, resulting in wasted memory requirements.

<u>Do binary spikes still make sense in NeuroAI?</u>

Sparsity is crucial. Neuronal spikes exploiting temporal sparsity are essential binary in nature (i.e. to spike or not). The operational principle has, to a large extent, constrained neuromorphic computation. It is possible to extend the notion of a spike to that of a packet. This view, while preserving sparse asynchronous information communication, relaxes the constraint to allow multibit data shuttling beyond simple rate coding. This perspective could foster opportunities for practical translations of high-precision AI models to neuromorphic hardware such as chip multiprocessor systems composed of neural processing units mediated by network-on-chip routers. Other coding schemes such as rank order coding, may still preserve binary spike paradigm while emphasizing weighting of the order in which spike arrives (simply put spread bit significance in time). It is also possible to use non-spiking encoding schemes such as biologically motivated oscillatory dynamics and phase coding schemes that are observed in single neuron and neural populations. However, given the dominance of DRAM memory, any scheme that results in unpredictable memory access will suffer penalties, so it is possible that predictable (e.g. layer by layer) synchronous updates of network state may continue to provide the best tradeoff of cost/throughput/efficiency.

<u>What organizing principles of neural computation (in Carver Mead sense) are we missing out exploiting?</u>

Future research should focus on algorithmic and hardware emulations of underexplored neural processes such as dynamic synaptic states, dendritic computation, astrocytic and glial cell mediated processes. On the hardware front, tight monolithic integration of these processes is crucial, e.g. minimizing the area footprint of analog VLSI implements of dendritic trees in silicon, or, more practically from a cost perspective, of finding methods to time-multiplex such concepts over shared processing elements. Promising ongoing efforts on new materials such as multi-gate ferroelectric field-effect transistors (FeFETs) for implementing rank order dendritic computation are being made. We advocate for engineering compact stateful synapse circuits to enrich spatiotemporal neural processing that transcend the point neurons used commonly today.

<u>What are the components of biology required for AI performance? Can developmental neurobiology teach us useful lessons for AI looking forwards?</u>

Concepts like structural plasticity and hippocampal neurogenesis are possible remedies for catastrophic forgetting. Neuromodulation as it pertains to continual learning can also offer promising avenues with an emphasis of dynamic adaptation across a wide range of timescales. Concepts like "fast weights" that store transient memories of the recent past in a Hopfield-network like associative memory have re-emerged as neurally plausible alternatives to attention units in recurrent neural networks. Other concepts like rapid declarative learning (i.e. the ability to learn facts and concepts quickly to the point of articulation), curriculum learning (i.e. learning from a set of formal pre-planned pedagogy), and one-shot negative (to avoid repeating crucial mistakes) are of outmost importance. Finally, opportunities for more direct human-in-the-loop learning could potentially improve artificial reasoning on tasks 'in the wild'.

<u>What current enthusiasm can enable future AI?</u>

Worldwide smartphone sales are greater than $1.2B per year and there are 9B smartphones currently in use. The worldwide competitive market means that smartphone Neural Processing Units (NPUs) are at the pinnacle of battery-powered AI processing technology in terms of cost, power, and throughput. It therefore makes sense to use these NPUs for other application areas like robotics, IoT, and prosthetics.

<u>What device and memory technologies are most crucial to enable future AI energy, throughput, and area improvements?</u>

Accessibility of DRAM interface technology to academia is a major opportunity to scale AI model development efforts. Co-design of hardware and algorithms have been stifled by memory capacity. Neuromorphic accelerator chips of today, while being massively parallel, are still limited to small-scale recognition tasks due to the convenient overreliance of SRAM cells. As parameter counts of state-of-the-art models scale to the billions, model translation to SRAM memories won't cut it due to area constraints (SRAM bits cost >20x DRAM bits). Making available commensurate memory capacity that is easily achieved with DRAM technology will be crucial for hardware translation of large models of arbitrary size at about 1/100 the cost of SRAM. Concerns about the cost of DRAM IP were raised. DRAM interface IPs are complex, proprietary and expensive and will require joint open-source efforts by funding agencies like the NSF and industry players. It is worth mentioning that DRAM memory achieves high capacity in 3D monolithic technologies that stack more than 200 layers. Tightly integrating compute and communication along with memory in this 3D framework coupled with the adoption of bit-level, modular and temporal sparsity enforcing algorithms will ensure area and energy footprints are minimized.

<u>What will be the main application areas or market sectors that can drive developments?</u>

- AI hardware accelerators
- Implantable Medical devices
- Brain connectomes in neuromorphic hardware

## Outlook:

We next present a Wishlist/ projection of short, mid and far term opportunities to advance neuromorphic research in the context of engineering efficient AI and other notable translational opportunities. Keeping in mind that 5-year aims may be overestimates while 10- to 20-year time horizons might be underestimates, as evidenced by decade-old projects like the Human Genome Project and the Brain Research Through Advancing Neurotechnologies (BRAIN) initiative. Successes chalked largely due to extensive collaborations between scientists and engineers, which this NeuroAI initiative seeks to foster.

<u>5-year:</u>

- Neuromorphic twin from fully-mapped mouse connectome
- Incorporation of more biological neurons, synapses and modulatory computation into AI
- Integration of 3D memories supporting sparse workloads
- Open memory and sensors (IP and fab support) to enable research and scaling for academics
    - DRAM chip interface IP blocks will allow academic DNN scaling
    - FeFET, RRAM, PCM & MRAM will enable novel CIM and PIM
- Neuromorphic deep brain stimulation technology
- Wearable technologies with neuromorphic computing (glasses, earables, health monitors.)
    - foveated projections, generative event-based/low-power AR/XR
- AI-driven programming of micro-code generation for neuromorphic computing

- o Open powerful smartphone NPUs to application outside smartphone companies
- o Hardware RNNs and SSMs as community chip developments
- o Enable current neural mean-field digital twins to be simulated in clinical times
- o Continual post training fine tuning, e.g. in robotics, for CoT and adaptive load and terrain, manipulation with increasing dexterity
- o Fleet learning will learn faster than single devices
- o Complexity gains in AI algorithms with more powerful neuromorphic primitives
- o Rewarding software with hardware metrics + AI-Coders
- o Personalized interdisciplinary education in neuromorphic engineering using AI
- o Accessible neural AI software / hardware for research and education

<u>10-year:</u>

- o Neuromorphic twin from fully mapped monkey connectome
- o Bio-compatible neuromorphic interfaces
- o Wearable neuromorphic sensors
- o Abstracted circuit-level neural twins for clinical decisions and therapies
- o Production of heterogeneous architectures that exploit all forms of sparsity
- o 3D integration of a diversity of distributed sensors, memory and compute
- o High-performance wearables that demand cost and power efficiency
- o Fully probabilistic AI
- o Resource-constrained uncertainty quantification: e.g., autonomous driving
- o Accessible testbeds for embedded AI for research and education

<u>20+-year:</u>

- o Neuromorphic twin from fully mapped functional human brain connectome and peripheral nervous system.
- o Closed-loop real-time high-resolution neural activity sensing, computing and actuation
- o Neuromorphic twins for new approach methodology for medical research and therapy
- o Algorithms / hardware that mimic the developmental computational stages to support efficient scaling and learning
- o Neuromorphic prosthetics (top-down or bottom-up approach using circuits or engineered neural tissue)
- o <1 kilowatt AI supercomputer
- o Human-level touch and olfactory sensing and low-latency feedback
- o Emerging technologies and neuromorphic chips in mass production
- o 50th+ year anniversary of Telluride Workshop

# Acknowledgments

This manuscript is adapted from an NSF Workshop report entitled "NeuroAI and Beyond" which was held on August 27-29, 2025 in Arlington, Virginia (Organizers: Cauwenberghs, Fellous, Fermüller, Sandamisrkaya and Sejnowski). The goal of the workshop was to explore future directions for NeuroAI. The organizers and workshop participants are grateful to Drs. Hillel Chiel, Barbara Oakley, Chiara Bartolozzi, Blake Richards and Tobi Delbruck for leading the discussions of their working group and for their contributions to the report. We also thank Xandra Dvornikova (Institute for Neural Computation, UCSD) for all the administrative help and to Drs. Soo-Siang Lim (NSF/SBE), Elizabeth Chua (NSF/SBE) and Grace Hwang (NINDS/BRAIN) for helpful discussions throughout this adventure. This material is based upon work supported by the National Science Foundation under Award No 2533276 (TS and JMF).

# Appendix

## 1. Workshop Participants

Embodied Cognition and Computation

*Hillel Chiel (CWRU)*
Jean-Marc Fellous (UCSD)
Steve Zucker (Yale)
Joe Paton (Champalimaud)
Alexander Ororbia (RIT)
Bing Brunton (UW)
John Yiannis Aloimonos (UMD)
Michael Bennington (CMU) - Scribe

Language and Communication

*Barbara Oakley (Oakland University)*
Christina Koppel (Imperial College London)
Xiaoqin Wang (JHU)
Evelina Fedorenko (MIT)
Cornelia Fermuller (UMD)
Shihab Shamma (UMD)
Seong Jong Yoo (UMD) – Scribe

Robotics

*Chiara Bartolozzi (Fondazione Instituto Italiano
    di Tecnologia)*
Carmen Amo Alonso (Stanford)
Yulia Sandamirskaya (Zurich University of
    Applied Sciences)
John Doyle (Caltech)

Christopher Kanan (Rochester)
Matthew Jacobsen (UMD) - Scribe

Learning in Humans and Machines

*Blake Richards (McGill University)*
Tony Zador (CSHL)
Terrence Sejnowski (Salk/UCSD)
Yann LeCun (NYU)
Michael Berry (Princeton)
Mike Stryker (UCSF)
Ali Minai (U Cincinnati)
Colin Bredenberg (Montreal) - Scribe

Neuromorphic AI Engineering

*Tobi Delbruck (INI)*
Brad Aimone (Sandia)
Ralph Etienne-Cummings (JHU)
Gert Cauwenberghs (UCSD)
Abhronil Sengupta (Penn State)
Jason Eshraghian (UC Santa Cruz)
Gina Adam (GWU)
Akwasi Akwaboah (JHU) - Scribe

Additional Student Participants:

Yianni Karabatis (UMD)
Eadom Dessalene (UMD)

## 2. SWOT Analysis Summary (all groups)

Compiled by Prof. Yulia Sandamirskaya (ZHAW, Switzerland).

Strengths

- NeuroAI facilitates multi-disciplinary research
  - We have a collaborative community (workshops, interdisciplinary grants, co-mentorship)
  - We can use multiple sources of funding (NSF, NIH, DoD, foundations, industry)
  - We enable multi-disciplinary exchange and in-depth collaboration, breaking disciplinary barriers
  - Strong interdisciplinary expertise within the NeuroAI community
- Neuroscience inspiration is foundational and central to further development of AI

- We can use wealth of data from neuroscience: connectomes, cell types, large-scale electrophysiology, behavior tracking in natural behaviors
- A growing corpus of high-quality electrophysiological, biomechanical, multi-omics, and behavioral data relevant to the development of embodied models, computational and physical
- We can use different models of neural computation, e.g. computation in different brain regions, such as declarative hippocampal learning system vs. procedural basal ganglia learning systems, cortical layers and thalamo-cortical loops, cerebellum architecture, etc.
- There are many unsolved problems in AI (alignment, energy, compute, data) where neuroscience and cognitive science can provide solutions
- NeuroAI is a driver for physical, embodied AI
  - NeuroAI focuses on connection to the physical world (embodied AI, robotics, sensing – key technologies for the future) and computational efficiency of AI (a key bottleneck and enabler)
  - Current advances in computational tools and frameworks are making it possible to perform full system simulations of animal behaviors
  - Developments in soft robotics and active materials set the stage for physical realizations of embodied computation in robotic systems
- Exciting field, attracting best student talent
  - Previous generation of NeuroAI (today's AI) has shown its potential to transform the world
  - Many jobs in AI today, especially deep learning, compared to pre-2016 in both academia and industry
- Grounded in strong foundational work from past decades
- Giving back to natural sciences
  - Enormous interest in the neuroscience and cognitive science community regarding how AI can inform new discoveries in their fields

<u>Weaknesses</u>

- Lacking theory: theorems, proofs of bounds and guarantees of NeuroAI
- Complexity of the field: interpretation of science and valuing scientific research is challenging
  - Many neurobiological components are very challenging to simulate or mimic (e.g. neuromodulators, peptides, gap junctions)
  - High-level cognitive capacities, e.g. emotions and reasoning capabilities, are difficult to integrate in neural models
  - Over-loaded terminology, a lot of jargon. It is a young discipline (needs to assert itself)
  - There is a lack of consistency and unification between the different simulation and modeling frameworks (akin to URDF for robotic modeling), which makes adopting new models or modifying existing ones to build new systems difficult
- Without consistency between these frameworks in how they interface with data, the new experimental data may remain inaccessible to modelers
  - Data sources are spread between many different labs and institutions without methods for centralized access or database searching
- Inaccessibility of cutting-edge hardware technologies and tools for academics to implement and test NeuroAI ideas. Too expensive
- Resistance from the mainstream AI
  - AI has much to learn from neuroscience to improve many aspects, but most AI researchers in industry and academia do not care about neuroscience
  - Mainstream AI companies prefer investing more in the current methods than exploring alternatives
  - Most NeuroAI researchers are academics who cannot compete with industry in resources
- Better training is needed to progress in the field
  - Most AI researchers are trained in computer science departments, lacking interdisciplinary insight (neuroscience, cognitive science, natural sciences, some parts of mathematics) and often lacking scientific and theoretical training to move the frontiers of AI.
  - The curriculum of computer science programs was not designed for teaching modern AI
  - Being productive in NeuroAI requires a lot of interdisciplinary knowledge and working in teams, rather than a single student working with a PI
  - There are not enough Specific NeuroAI training opportunities for postdocs and grad students

- o Better ethics training is needed
- Not enough 'success stories' (or not visible enough)

<u>Opportunities</u>

- New areas of applications for NeuroAI
  - o Biomedicine/Healthcare, Neuroscience, Physics, Robotics, Space
  - o Edge NeuroAI can be expanded for personalized and untethered applications
- Identify specific gaps in current AI paradigms and focus on biologically-inspired methods to address them
  - o Explore biologically-inspired paradigms that are radically different than the currently dominant ones.
  - o Potential to conceive of next-gen AI to leapfrog current LLMs
  - o Genuine creative and new thinking about AI, beyond incremental/ technical advancements
  - o Need to explore all biological properties (different neurons, glia, biochemistry, etc.) and computation (oscillation, attention, microcircuits, cortical-subcortical interactions, etc.) of the nervous system beyond matrix-vector multiplication and rectification for AI models
- Well positioned to integrate substantial knowledge from neuroscience, cognitive science, ethology, computing, robotics, dynamics & control, and AI — these are individually mature fields
- *In vivo - in silico* computational hybrids, BCI
  - o Advances in "dynamic databases" and *in silico* experimental platforms using existing datasets could alleviate (but not replace) some of the reliance on animal experiments and provide opportunities for massively parallelized investigations
- Alternative materials and fabrication technics for computation. Inspiration for future computing hardware
- We need to create degree programs aimed at training students in AI, which has grown beyond computer science. There are a lot of opportunities to do things better if built from the ground up
- Use NeuroAI as a massive springboard for K-12 teacher training, potentially by working with the new National Academy for AI Instruction (springing from Microsoft, OpenAI, and the American Federation of Teachers), and other online training programs
  - o Embodied NeuroAI could be used in educational settings to facilitate child- computer interactions that have a strong neuroscientific basis
- We need better engagement with and training for policy makers, and scientific representation in government
- Because it is based on neuroscience principles, NeuroAI will likely offer new hypotheses about how the brain is organized and works and these hypotheses can be tested experimentally by neuroscientists

<u>Threats</u>

- Academia can't compete with industry for resources (human and tech), making it hard to seed research long-term, beyond the next quarter's profits
  - o Current AI paradigms may come to dominate very rapidly and leave no space for NeuroAI
  - o Divergence of resources from sustainable, incremental research to chase impossible, overpromised goals set by tech giants and big labs/projects – both in AI development and understanding the brain
- Missing graduate students, professors, postdocs to develop high risk ideas and educate future AI practitioners
  - o NeuroAI could disappear if there is no critical mass of young talents to sustain it
  - o Relatively few students trained with the right combination of skills in neuro, computation, and engineering
  - o Better interdisciplinary undergraduate education is needed
- Lack of collaborative projects between multiple labs due to lacking incentives and funding instruments
  - o Continued development of models, tools, and theories without careful attention to enable exchange and alignment between approaches can lead to divergence of understanding as different groups become more cloistered in their preferred framework
- AI – both neuro- and deep learning-based -- could be misused for spreading wrong information and noise, privacy and security threats both digital and physical AI
  - o Ethical and legal considerations should be developed to ensure NeuroAI is not inadvertently or intentionally misused

### 3. SWOT Analysis (Students and Postdocs)

Strengths

- Network style grants (multi-institution, multi-lab funding opportunities)
- Training programs for young scientists
  - Telluride program
  - Neuromatch
  - NeuroPAC
  - Student fellowships (gives independence)
- Interdisciplinary research (bridging theory-experiment gap)

Weaknesses

- Siloed academic departments and research labs
  - Difficulties taking classes & forming collaborations across departments
- Ethics/Safety education and institutional/organizational resources
  - Foresight into mitigating risks of tech development (history of prev. failures)
  - Standardization of principles regarding ethics and use of AI tech
  - Evaluating ethics/safety contributions in grant evaluations or hiring decisions
- Public outreach, public feedback regarding tech development

Opportunities

- Funding opportunities for lab exchanges
- Competitions: incentivizes improvement & collaboration with independence and creativity
- Shared Resources
  - Policy that gives incentives to industry for establishing shared compute
- NSF-Industry HPC clusters
  - GPU and AI accelerators are revamped frequently to meet AI demands
  - What happens to 'old' compute?
  - Could there be chain of resource flow from industry to academia?
  - Long-term incentives for both
- Academia gets to work on heavy duty model training
- Industry gets to fetch from a pool of talent that are familiar with their resources
  - Bring MOSIS back! (Will advance Chip design talent)

Threats

- Funding uncertainty
- Visa and immigration issues for conferences, internships, research
- Industry / international brain drain (domestic students)
- Postdoc compensation

**4. Workshop Schedule**

# NeuroAI and Beyond

| Wednesday, August 27 | |
|---|---|
| | |
| 6:00 PM – 6:30 PM | Reception |
| 6:30 PM -7:00 PM | Opening Remarks by NSF and Workshop Goals |
| 7:00 PM – 8:00 PM | *Keynote address*:  Carmen Amo Alonso<br>          "NeuroAI Meets Control Theory" |
| | |
| **Thursday, August 28** | |
| | |
| 9:00 AM - 12:00 PM | Opening remarks Dr. K. Husbands Fealing, NSF/SBE |
| | ***Initial Position Statements*** |
| 9:00 AM - 9:30 AM | Embodied Cognition and Computation |
| 9:30 AM - 10:00 AM | Language and Communication |
| 10:00 AM - 10:30 AM | Robotics |
| 10:30 AM – 11:00 AM | *Break* |
| | |
| 11:00 AM – 11:30 AM | Learning in Humans and Machines |
| 11:30 AM – 12:00 PM | Neuromorphic AI Engineering |
| | |
| 12:00 PM – 1:30 PM | *Lunch* |
| | |
| 1:30 PM – 3:30 PM | ***Breakout Session 1*** |
| | |
| 3:30 PM – 5:10 PM | ***Interim Position Statements*** |
| | |
| 3:30 PM – 3:50 PM | Embodied Cognition and Computation |
| 3:50 PM - 4:10 PM | Language and Communication |
| 4:10 PM - 4:30 PM | Robotics |
| 4:30 PM - 4:50 PM | Learning in Humans and Machines |
| 4:50 PM- 5:10 PM | Neuromorphic AI Engineering |
| | |
| 5:10 PM | Training and Workforce Development |
| | |
| 5:30 pm | Working Dinner |
| | |
| | |

| Friday, August 29 | |
|---|---|
| | |
| 9:00 AM – 10:30 AM | ***Breakout Session 2*** |
| | |
| 10:30 AM – 12:10 PM | ***Interim Position Statements*** |
| | |
| 10:30 AM – 10:50 AM | Embodied Cognition and Computation |
| 10:50 AM - 11:10 AM | Language and Communication |
| 11:10 AM - 11:30 AM | Robotics |
| 11:30 AM - 11:50 AM | Learning in Humans and Machines |
| 11:50 AM- 12:10 PM | Neuromorphic AI Engineering |
| | |
| 12:10 PM – 1:30 PM | *Lunch* |
| | |
| 1:30 PM – 3:00 PM | ***Breakout Session 3*** |
| | |
| 3:00 PM – 4:00 PM | *Closing Discussion* |
| 4:00 PM | *Farewell* |

## 5. Personal Statements from the Participants

<u>Gina Adam</u>

NeuroAI will benefit significantly, and in some applications, probably require new types of neuromorphic hardware technologies that can interface with sensors and actuators in an efficient way. The theme of embodied AI came up time and time again during the Workshop discussions, which was intriguing to me because it would be a prime example of the need for co-design across the entire stack like our team proposed. The issue of neuromorphic co-design is a challenging one because it should be done in a cohesive way, without stifling innovation in the individual levels of neuroscience, algorithms, hardware and sensors. One of my main concerns is the lack of research and educational infrastructure to achieve such interdisciplinary work. What platforms can be put in place for researchers and students to be able to explore integrated neuromorphic computing and sensing hardware in an accessible way? Unfortunately, the current cost of new types of hardware can be prohibitive for many research groups and there are too few opportunities to support experimental integration of neuromorphic computing with sensing and actuation towards embodied AI. Hopefully, we as a community can work together to define some initial research goals and the infrastructure we would need to work towards neuromorphic integration and embodied AI.

*Gina C. Adam, Ph.D., is an Associate Professor of Electrical & Computer Engineering at George Washington University. Her research interests include novel memristive synaptic nanodevices and neuronal circuits and their integration into emerging hardware paradigms for neuro-inspired computing.*

<u>Brad Aimone</u>

The days of weak brain-inspiration are over. For too long it has been possible for people to take their toy ideas of how cognition works and find some justification in the underconstrained complexity of the brain. That isn't going to work anymore - we now have exascale supercomputers and brain-scale neuromorphic systems and thanks to BRAIN we now have full-scale connectomes and more sophisticated measurements of spatial and temporal neural dynamics. As such, we can start to falsify the bad ideas and throw them out and we can start moving forward as a field. We can and should develop algorithms and frameworks that embrace, rather than avoid, the brain's heterogeneity, parallelism, stochasticity, and spatio-temporal complexity. True brain-inspired algorithms are right around the corner, and this will not only help AI and numerical computing broadly, but it will loop back and help us finally understand how the brain works.

*Brad Aimone, PhD, is a Distinguished Member of Technical Staff in the Neural Exploration and Research Laboratory at the Center for Computing Research at Sandia National Laboratories. His research sits at intersection of neuromorphic computing and theoretical neuroscience, with a particular focus on how insights from the brain can be used to influence future scientific computing, NeuroAI, and emerging hardware technologies.*

<u>Akwasi Akwaboah</u>

I am convinced a coupled hierarchical space-time perspective (be it diffusive, wave-like or cellular automata-like) to modelling intelligent systems will optimally leverage neuromorphic technology. The continuous dynamics of diffusion and wave propagation have been revealed in cortical dynamics as studied in recent neuroscience research. These may well be the underlying modality of perception and cognition. In a similar measure, neuromorphs have and continue to engineer systems like retinomorphic resistive grids and silicon cochleae that are inherently spatiotemporal, achieved beautifully in the fabric of analog VLSI. In contrast, the development of modern AI models have been inordinately fixated on discretized spatial and sequence processing with a current obsession for tokenization. Tokenization, while being effective, does not explicitly connote temporal association and thus may not be the Occam's razor of intelligence. Rather, I posit that hardware-friendly learning and inference methods that commensurately assign spatial and temporal credits could reveal ultimate insights to artificial perception and reasoning. A balanced space-time view of intelligence, which has been achieved by the biological brain over the billion-year evolutionary process, stands to offer efficient solutions that amortize computational and energy costs in four physical dimensions. Let's see what the future brings this early-career neuromorph!

*Akwasi Akwaboah is a PhD Candidate at the Department of Electrical and Computer Engineering, Johns Hopkins University. He was a NeuroPAC visiting scholar at the Integrated Systems Neuroengineering lab, University of California San Diego in the fall of 2024. His dissertation, by means of mixed-mode VLSI and other in-silico methods, seeks to bridge the hardware and algorithmic divides between artificial and biological intelligences. His current research interests span the following domains: visual psychophysics, computer vision, spatiotemporal sparsities, neural interfaces and neuromorphic systems.*

<u>Yiannis Aloimonos - Embodiment in the service of space and action (space-time)</u>

Embodiment and its consideration change the kinds of questions that may be asked about a cognitive system. In computer vision for example, if a system with vision is moving in its environment, the questions asked are about finding the ego-motion and segmenting the scene. But when biological systems move, their brains tell their vision systems about the movement they are generating. In this case, equipped with this knowledge, we can ask new questions, like: given the current image can we predict the next one? Consideration of embodiment introduces a large variety of new interesting problems. Following such ideas, my research group (prg.cs.umd.edu) has been investigating (a) sensorimotor representations of visual space and (b) representations of perceived human action that can be used in teaching robots new tasks. Given (a) and (b) we investigate (c) embodied language processing.

(a) **Embodied Visuomotor Scale**: Today's robots for the most part employ 3D sensors, i.e. cameras that provide the exact distance of any object in the scene, in meters or feet. A consequence of this is that all robots need to be calibrated. But biological systems do not work in this way. For example, as you read these words, in your immediate environment you can see many objects and although you don't know their metric distance, you have intimate knowledge of exactly where they are, because you can put your finger at any point of those objects. Thus, visual space could be encoded in motor space – we demonstrate this by building robotic mechanisms that can decide to go through a hole (a door) without knowing the opening's absolute size or jump over a gap without knowing the gap's absolute size.

(b) **The Grammar of Action**: When we perform an action with our hands and tools, a fundamental concept is the one of contact. Our hands come into contact with tools and objects, or the tools come into contact with objects. Using those contacts in a video depicting human manipulation action, we can basically segment the video into subvideos containing individual primitive actions. This is a form of tokenization with each token representing a primitive. Because we have two hands, this is a hierarchical tokenization and we can now employ modern techniques to develop robot foundation models or behavioral foundation models, where given a video of an action we can learn to predict the next behavioral token. This is ChatGPT for video – where video has become a sequence of tokens (reach, grasp, twist, push, pull ...).

(c) **Language Grounding**: When humans comprehend language, their motor and visual cortex get activated (verbs, tools). This is the grounding of language. This is not the case for today's LLMs because they understand words only in relationship to other words (trained using text corpora). To create alignment we need robots and humans to understand language in a similar way. This would require the development of the PRAXICON (πραξις = praxis = action). The term praxicon was introduced by Liepman in 1908 in his work on apraxia, and it meant motor representations for many actions that are stored in procedural memory. The time is ripe to develop a praxicon, i.e. motor representations (including tactile feeling) for the verbs in any language. Like the LEXICON (λεξις = leksis = word) was instrumental in achieving progress, the PRAXICON would allow embodied language processing making it possible to create a visuo-motoric simulation of a sentence of words, thus bringing Linguistics to new heights.

*Yiannis Aloimonos (prg.cs.umd.edu), PhD, is professor of Computer Science (CS) and of Electrical and Computer Engineering (ECE) at the University of Maryland, College Park and the director of the Computer Vision Laboratory, with appointments in the Institute for Advanced Computer Studies (UMIACS), the Institute for Systems Research (ISR) and the Maryland Robotics Center (MRC). His research interests include Active Perception and the modeling of vision as an active, dynamic process for real time robotic systems. For the past several years he has been working on bridging signals and symbols, specifically on the relationship of vision to reasoning, action and language with applications to imitation learning in robotics and the grounding of natural language.*

## Carmen Amo Alonso - Control architectures are fundamental in learning systems

The rise and success of deep learning across different disciplines and domains has been truly impressive. Something else that is impressive, yet often understated, is how far a principled understanding of dynamical systems and our ability to control them has taken us. From aviation to the internet, control is at the heart of what makes these systems reliable, scalable, deployable, and ultimately trustworthy in our society. We have some evidence of how dynamics and control mechanisms spontaneously emerge from training under the current paradigm and are present in existing learning architectures. We are also starting to understand the power of directly designing control mechanisms for these architectures. What we are currently missing is a more principled approach to these processes using tools from control theory and dynamical systems. An understanding of the system's dynamics will enable us to design learning systems in a more principled way, leading to more robust and efficient designs as well as better understanding of possible failure modes. Furthermore, a systems architecture view of the full learning system, from abstractions and memory to embodiment, could enable us to exploit physical constraints in the full system design. These "constraints that deconstrain" are fundamental to most scalable, robust, reliable, and efficient systems that we have, both in nature and in engineering. I believe that it is essential to design the full stack of a learning system, from abstractions to its lowest-level physical embodied hardware, using systems architecture theory. Machine learning has achieved a multitude of successes, but true artificial intelligence will only be reached once we understand (and take advantage of) how the full system architecture works together. In turn, this theory will also allow for a better understanding of natural intelligent systems as well.

*Carmen Alonso is a Schmidt Science Fellow at Stanford University. She works on the foundations of Artificial Intelligence, particularly for language applications. Drawing from her background in aerospace engineering (B.Sc. Technical University of Madrid, 2016, M.Sc. Caltech 2017), as well as in control theory (Ph.D. Caltech 2023), her work aims to adapt the mathematical principles of control theory--key in enabling safety in the aerospace industry--to address the unpredictability and risks associated with AI technologies, as well as to create more principled and efficient AI models.*

## Chiara Bartolozzi

Modern AI is showing impressive performance in some tasks that we consider typical of human intelligence. The most successful AI systems are foundational model, that replicate language production and interpret sensory data, such as videos and audio recordings. They do so by using some of the computational principles observed by neuroscientists, but it ignores the most fundamental aspect that human (and animal) brains are not disembodied entities that live in an abstract cloud. Rather, biological brain computation is entangled with how it receives information through the body, physically filtered and pre-processed by how sensors interface with the external world, and how information is shaped by the actions of the body. There are continuous sensorimotor feedback loops at different spatial and temporal scales that process information, integrate predictions and generate actions. While language is one of the landmarks of human intelligence, we can argue that it stems from a different, or more fundamental, type of intelligence, that is being able to generate physical actions with concrete consequences in the world. Changing the perspective on how we look at intelligence, as embodied and with a specific (action directed) goal, might steer the research towards solutions that work more efficiently, using less computational resources and less data, affordable by independent researchers, hence, more democratic and freer.

The biological brain architecture has been shaped by evolution and development and is continuously refined by learning during the individual's lifetime. While foundational models could deliver such an infrastructure, it is still lacking a lot of the aspects of biological brains, as in the massive feedback towards primary sensory areas, in the diversity of brain areas (cortical and sub-cortical), of neuron models, of bodies and sensors, and in the nested multiscale sensorimotor control loops.

Grounding language models in an architecture that has been built with embodiment and actions-based intelligence and that better incorporates information from neuroscience might generate abstract reasoning and language that is grounded on an intuitive grasp of the physical world and can generate through understanding, rather than through mirroring by fitting data statistics.

*Chiara Bartolozzi, PhD, is Senior Tenured researcher at the Italian Institute of Technology, where she coordinates the Event-Driven Perception for Robotics research unit. Her research interests include the application of neuromorphic technologies, from sensing to actuation, to robots, embodiment, development of neuromorphic sensors and edge applications.*

## Michael J. Bennington - Rigid and soft embodiment

The mechanics and functional anatomy of the body are integral to understanding how biological systems behave and interact with their environment. The body serves as the literal bridge between the nervous system and the outside world, containing the structures and actuators (e.g., muscles) that allow the nervous system to move around and interact with the world and the sensors that provide feedback to it. The body can also allow for lower-level control to be computed outside of the central nervous system (e.g., reflexes and preflexes). This embodiment of the nervous system seems critical to the emergence of intelligence in biological systems. Therefore, when developing a more general theory of embodiment, we need to understand the functional role of the system's biomechanical components, as well as how they change across species, size, evolutionary time, and body plans. Mathematical tools and functional anatomy frameworks exist for doing this in rigid-body systems (i.e., animals with an endo- or exoskeletal system). This problem becomes more challenging with soft-bodied systems in which actuators, passive components, and structural elements are often collocated in the same anatomical structures. These systems can perform a hugely different set of motions, actions, and behaviors from their rigid body counterparts. Therefore, we may need different "praxicons" and theories for the emergence of these praxicons from the underlying functional anatomy for rigid- and soft-bodied systems. In seeing where these very different mechanical systems differ, and where they overlap, we may be able to gain a deeper understanding of the connection between the brain and the body.

*Michael J. Bennington is a PhD candidate in Mechanical Engineering at Carnegie Mellon University. His research interests include the biomechanical modeling and control of muscular hydrostats, the role of mechanical intelligence in behavior, and methods for capturing animal-to-animal variability.*

## Michael Berry

In my first 20 years as a neuroscientist, I used multi-electrode arrays to record from populations of retinal ganglion cells to study how they process and encode visual information. This work has given me extensive experience building computational models to explain circuit-level mechanisms as well as analyzing population neural codes using information theory and other probabilistic models, like the maximum entropy and hidden Markov models. But in my 20+ years of exploring the computations carried out in the retina, what has most struck me and captured my imagination is the ability of the retina to make predictions about the upcoming visual stimulus. The fact that a circuit as simple as the retina can carry out such sophisticated computations has led me to hypothesize that predictive computation might also be a feature of neocortical circuits.

In this vein, I have begun to study neural computation and coding in the visual cortex of mice, using 2-photon calcium imaging. Here, we have presented mice with repeated sequences of images and movie clips arranged into temporal sequences, to see how temporal context effects the neural code. We are formulating network models to explore the circuit mechanisms underlying these phenomena, as well as to explore novel ideas about the role of predictive computation.

Upon studying the neocortex, I have been drawn to the fact that it operates via powerful reciprocal loops with thalamus, basal ganglia and cerebellum as well as connections up and down the cortical hierarchy. An approach to understanding these functional dynamics that I believe holds promise is identifying key computations entailed by the biological details of microcircuits but assuming that they are roughly repeated across brain areas. As a first step in this agenda, I have formulated a model of thalamocortical loops, which is involved in evidence accumulation and motor planning.

The perspective of repeated microcircuits promises to provide insights for AI algorithms and neuromorphic engineering by identifying key computations in the brain that are: 1) useful if they are repeated across the hierarchy, and 2) may not be included in current algorithms and devices.

*Michael J. Berry II, is a professor of neuroscience at Princeton University. He has studied neural computation and coding in the retina and primary visual cortex.*

<u>Colin Bredenberg</u>

Artificial intelligence and machine learning provide natural frameworks for describing the computations implemented by internal brain dynamics and their connections to a wide array of complex behavioral phenomena, ranging from low-level perception and action to cognition and language. Further, statistical data science methodologies have proved to be an enormously useful tool for the neurosciences for the purposes of experimental design, quantifying naturalistic behavior, and making sense of high-dimensional neural data. As such, these methodologies are rapidly becoming indispensable for modern systems and cognitive neuroscience.

Conversely, neuroscience continues to play a major role in inspiring the development of novel artificial intelligence algorithms, and NeuroAI scientists often provide valuable perspectives for improving upon the limitations of existing AI systems. Modern AI algorithms are generally treated as inscrutable 'black boxes,' and are tested on superficial performance metrics that are susceptible to overfitting specific datasets and hardware requirements. As a consequence, these algorithms tend to suffer critical failures on deployment that are difficult to diagnose and can be difficult to extend to solve novel problems. Because of their focus on neural representations, reverse-engineering, more fine-grained analyses of intelligent behavior, and people themselves, trained NeuroAI scientists continue to provide insights that make modern algorithms more robust, human-like, and better at interfacing with human users.

Therefore, NeuroAI scientists continue to make valuable contributions to both neuroscience and artificial intelligence research. However, for this virtuous cycle to continue, a clear-eyed assessment of the limitations of the contemporary state of the NeuroAI discipline is necessary. First, there is a powerful need for scientists who are equally well-versed in neuroscientific methods as they are in artificial intelligence methods. Interdisciplinary education of this kind is still inaccessible at the undergraduate level at many universities, and near-future curriculum reform will become increasingly necessary as the amount of knowledge required to contribute to cutting-edge research increases. Second, given the ascendancy of industrial-scale artificial intelligence development, there are strong financial pressures that concentrate the attentions of NeuroAI and AI researchers alike into subfields that could provide immediate benefits to AI companies (natural language processing, computer vision, robotics), at the expense of subfields that show promise for medical healthcare or longer-term technology development (olfaction, somatosensation, emotional processing, synaptic plasticity, microcircuit-level computation, brain-computer interface design, development and self-assembly, etc.). If insufficient effort is invested into supporting NeuroAI as a legitimate and independent subfield of research, the field risks being heavily biased by inappropriate tools imported from artificial intelligence research, developed under large-scale computing constraints and profit requirements, without appropriate respect for the genuinely differing realities of biological systems.

*Colin Bredenberg, Ph.D., is a postdoctoral researcher working with Professors Blake Richards and Guillaume Lajoie at the Université de Montréal and Mila – the Quebec AI Institute. He employs artificial intelligence techniques to investigate the neural mechanisms of learning in diverse domains, from perception to motor control and navigation.*

### Bing W. Brunton - The future of NeuroAI is embodied

To understand "intelligence," neuroscientists and AI researchers often make an abstraction that a thinking, behaving animal makes decisions as a point-mass, or better yet, exists in the realm of pure thought. This worldview neglects a simple truth: *that the brain and nervous system of animals evolved jointly with the body they inhabit*, so that coordinating the movement of this body is the ultimate point of neural function[1].

Indeed, the brain is approximately not connected to the external world except through the body. It receives no sensory inputs except those transduced through the body and can enact no consequences except through muscles and other actuators embedded in the body. Further, biomechanical features and constraints are critical in shaping neural function, and how animals move through the world determines how they can actively gather new sensory information when situated in an environment.

*Embodiment, then, is the concept that the function of the brain is inexorably shaped by the body*. It is the natural lens through which all neural function can be considered, and an idea that is increasingly gaining attention in NeuroAI. To embrace the embodied intelligence evident in living systems, models of the brains and advanced NeuroAI systems should consider three principles: *feedback*, *biomechanics*, and *modularity*. First, feedback is a ubiquitous characteristic of embodied systems. Biology is built on constant and multiscale feedback. The view of the brain as a passive computational engine that tries to form representations of the world without interacting with it denies us our essential agency: to act, to change, to leave the world a (hopefully) better place than we found it. Second, biomechanical features of a specific body matter in understanding neural control. Harnessing the mechanical intelligence of the body greatly simplifies the demands of nonlinear neural control. Third, modules compose the overall embodied system. Even as we advocate an integrative approach to neuroscience, it is undeniable that parts of the nervous and musculoskeletal systems can be reduced and characterized in isolation. Models of different modules can be made at different resolutions and learned from different datasets.

As a specific example of an embodied model that integrates comprehensive datasets from different modalities, we and collaborators are developing a fully integrated neuromechanical model of the fruit fly. The recent completion of comprehensive synaptic wiring diagrams, known as connectomes, of the fly brain and ventral nerve cord means it is now tractable to hook up computational models of neural dynamics at the resolution of cells and synapses to whole-animal biomechanical models in simulated physics engines[2]. With such platforms, we will soon be able to investigate active behavioral sequences, particularly to understand how specific recurrent neural circuits support robust behaviors, and how such neural implementations may inform design of autonomous synthetic agents.

The convergence of software tools, open datasets, and a culture of collaborative science makes now a good time to take this bold shift in NeuroAI towards embodiment. We stand to gain a comprehensive, holistic understanding of neuroscience and a promising path towards general intelligence. After all, the only examples of intelligent systems we agree on (i.e., some humans and animals) are embodied. Therefore, it is reasonable to hypothesize that developing embodied NeuroAI systems may be a necessary path towards developing intelligence. The brain sits not in a jar but should be situated where it belongs: inside a body it evolved to sense and control, surviving and thriving in an uncertain world.

*Bing W. Brunton, Ph.D., is a Professor of Biology and the Richard & Joan Komen University Chair at the University of Washington (UW) in Seattle, with additional affiliations at the eScience Institute for Data Science, the Paul G. Allen School of Computer Science & Eng., and the Dept. of Applied Math. Her research group develops data-intensive methods to build models of the nervous system and body that can interact with a changing environment and predict responses to unexpected manipulations, using approaches from dynamical systems & control, deep reinforcement learning, computer vision, and physics-constrained simulations.*

1. A version of these ideas was previously published as an essay in *The Transmitter*.
2. See our proof-of-principle study of fly connectome simulations that produce leg motor rhythms (Pugliese et al., 2025), which will interface with an extended version of a whole-body biomechanical model of the fly in MuJoCo (Vaxenburg et al., 2025).

## Gert Cauwenberghs

At this time of amazing developments in the burgeoning field of artificial intelligence seemingly with unlimited cognitive abilities it is important to be reminded of the gold standard of natural intelligence in biological neural systems which have evolved to perform robustly, effectively, and efficiently, all at once. Their extreme resilience to environmental and operating conditions and their extreme frugality in operating near fundamental thermodynamic limits of noise-energy efficiency are unsurpassed in any manmade systems to date. In contrast, the ferocious appetite of today's AI for energy consumption at unsustainable levels endangering our future, compounded by the brittleness of today's AI at even the simplest unmalicious challenges and natural real-world conditions, call for a refocused collective effort in applying neuromorphic principles to endow AI with neural natural intelligence. Meeting with experts and stakeholders at the workshop has invigorated discussions on how to accelerate these developments in a most effective and practical manner. A key realization shared by many at the workshop is the urgency of action and support of these and related efforts, especially considering substantial investments leading advances in AI abroad, some of which developed under cover with dubious intentions potentially for use against humanity, as well as declining interest in advancing science, technology, engineering and mathematics (STEM) by the general public in the United States. Continued dominance in AI and chips research and development in the United States requires substantial federal investment and incentives, which would be strongly supported by renewed and strengthened research programs in neuromorphic AI engineering. One of such opportunities for continued NSF support is the annual Telluride Workshop on Neuromorphic Cognition Engineering which entered its 30th year last year and which has brought forth several generations of leading researchers in the field in academia, industry, and government labs.

Relevant references: (Mead, 1989; Cauwenberghs, 2013; Churchland and Sejnowski, 2017; Wan et al., 2022; Mead, 2023)

*Gert Cauwenberghs, Ph.D., is Distinguished Professor of Bioengineering and Co-Director of the Institute for Neural Computation at UC San Diego. His research focuses on micropower integrated biomedical circuits, neuron-silicon and brain-machine interfaces, neuromorphic engineering, and adaptive intelligent systems. Under his leadership, the Integrated Systems Neuroengineering Laboratory has been delivering integrated circuit solutions at record noise-energy efficiency for neural interfaces, neuromorphic computing, and RF signal intelligence.*

Hillel J. Chiel - Foundations of embodied cognition

Adaptive behavior, behavior that allows animals to survive and reproduce, emerges from the coupled dynamics of the brain, body, and environment (Chiel and Beer, 1997). Embodiment is critical for understanding the mechanisms of adaptive behavior. Studying both the nervous system and the biomechanics of the body in an experimentally tractable system, the marine mollusk *Aplysia californica,* has clarified behaviors that may underlie our higher cognitive functions – flexibility and robustness (Lyttle et al., 2017), multifunctionality (Webster-Wood et al., 2020), rapid responses to changing environmental conditions (Gill and Chiel, 2020), and how *local* changes in synaptic connections can lead to *global* changes in behavior (Tam et al., 2020). Neuromechanical models clarify how these behaviors emerge from ongoing interactions of biomechanics and neural dynamics (Webster-Wood et al., 2020) and help create mathematical tools for understanding responses to perturbations (Wang et al., 2022). Biologically-inspired robots demonstrate that incorporating reflexes enables complex terrain locomotion without pre-computing responses (Espenschied et al., 1996), and that plasticity can enhance a soft grasper's effectiveness (Li et al., 2025). A framework for determining the key forces that dominate behavior as a function of size and speed (Sutton et al., 2023) provides constraints and opportunities for control. More generally, creating cognitive agents through the biological process of evolution, development and plasticity rather than through standard techniques of design, manufacture and linear control may be a way to generate truly novel autonomous agents that can adapt to and shape complex environments.

*Hillel J. Chiel is a Distinguished University Professor at Case Western Reserve University in the Departments of Biology, Neurosciences and Biomedical Engineering. His laboratory focuses on the mechanisms of adaptive behavior. The laboratory's primary focus has been on using electrophysiology, biomechanical analyses, and computer modeling to understand the mechanisms of feeding behavior in the marine mollusk Aplysia californica. In collaboration with colleagues, he creates theoretical models of adaptive behavior and uses them to develop tools for characterizing systems that must rapidly adjust to changing environmental conditions, to understand the general constraints imposed by the physics of the world on behaviors over many ranges of size and speed, and for developing biologically-inspired robots. He is a Fellow of the Institute of Physics (London).*

Tobi Delbruck

It turns out that slavishly imitating the brain was holding us back; digital continued to scale and was designable and manufacturable at very complex scales. Mixed-signal continues to be useful at the sensor periphery. I want to approach the next phase with a more open mind to practical aspects of commercialization, which progresses small steps at a time funded by on-going profits. I'm excited about the next few years where we will increasingly see neuromorphic sparse computing principles appearing in mass-production smartphone, wearable, and robotics hardware AI systems. It is also obvious to me (and I believe most of the attendees) that the current great progress in symbolic AI is just the tip of the iceberg. Why is it that many decades after the first robots, the most capable one we can buy for a household is a (rather dumb) vacuum cleaner puck? Clearly the abilities of robots to take over tasks in homes, workplaces, and agriculture are still a long way in the future, but it is a no-brainer to exploit the developments of smartphone NPUs (manufactured in >\$100M/yr) for robot brains, just as GPUs were hijacked for AI training and to work on the hard problems of robotic active manipulation in human environments and continuous fleet learning from experience.

*Tobi Delbruck (IEEE M'99–SM'06–F'13) earned a Ph.D. from Caltech in 1993. Since 1998, he has been a Professor of Physics and Electrical Engineering at ETH Zurich, leading a group focused on neuromorphic sensors, neural hardware, machine vision, and neural robotic control. He co-organizes the Telluride Neuromorphic Cognition Engineering workshop. Delbruck is a past Chair of the IEEE CAS Sensory Systems Technical Committee. He has worked in electronic imaging at Arithmos, Synaptics, National Semiconductor, and Foveon, and founded three spin-off companies. A recipient of 13 IEEE awards, he was named an IEEE Fellow for his contributions to neuromorphic sensors and processing.*

John Doyle – Robustness and layered architectures

For me, this meeting and document is a refreshing summary of the challenges in NeuroAI.  The acute challenges of robustness and architecture are familiar to me but in a newly challenging domain.  Modern optimal control by 1975 promised scalability, flexibility, and accessibility but failed miserably on robustness. Robust control expanded and largely replaced optimal control in most engineering application domains at the expense of accessibility. By 2000 robustness was also becoming emphasized in complex networks from bacteria thru the internet, and it became increasingly clear that extreme robustness requirements in networked systems required Universal Layered (ULA) aka Layered Control (LCA) Architectures that are ubiquitous but were poorly understood.

Since then, a rich theoretical (and software) infrastructure for network architectures has developed and is widely used in some domains, such as systems biology and power grid control.  Its application to neuroscience and AI is newer but promising (e.g. Carmen Alonso and Terry Sejnowski). The most important and accessible theories are Layering as Optimization (LAO), and System Level Synthesis (SLS) and the hourglass and bowtie motifs. SLS is the newest theory and addresses the greatest obstacle to LAO, which was that the lowest layer feedback control systems often had delays and sparsity that made the needed optimization the most infamous intractable problem in control theory. SLS completely changed this and made lower layer feedback control with delays, sparsity, and locality both scalable and robust.

This has opened a new theory frontier in both control and NeuroAI that is the most glaringly absent topic in this report.  A major challenge is to make this more broadly accessible and broaden applications beyond visual cortex, LLMs, and robotics.  A striking result is that the massive internal feedback from motor to the rest of the brain (e.g. LGN) being newly highlighted by experimentalists is necessary to overcome intrinsic delays and sparsity in neural hardware while delivering robust embodied performance.

A hopefully minor obstacle to the adoption of ULA in neuroscience is that much theory has previously been dominated by "new sciences" of complexity and networks, with chaos, criticality, fractals, scale-free, etc… These have been thoroughly debunked among experts in their original favorite application domains of cell biology, earthquakes, wildfires, and the internet, but remain inexplicably popular in neuroscience, though fortunately not in this group. We can now make precise how the coarse graining across levels of "emergence" is exactly the opposite of the fine tuning across layers for robustness in ULA, as illustrated by the human brain.

At best, neuroscience has drawn on the important but previously fragmented engineering areas of information, computation, and control, that ULA now unifies.  ULA+SLS is a rigorous and coherent theory clarifying otherwise vague connections between these frameworks as well as of hierarchy, modularity, teams, adversaries, and of course robustness.  Accessibility remains a major challenge, and I'm eager to focus my retirement on this and cancer immunology (where ULA is proving equally essential).  Fortunately, ULAs are everywhere so simple and accessible starting examples are Legos, clothing, and baking, and all around us in housing, cities, transportation, supply chains, manufacturing. The popular term "99% invisible" applies ubiquitously in our most sophisticated architectures, adding to the problems of accessibility.

Layers, levels, stages, and laws are all the essential elements of robust architecture, but only layers is unique to architecture as the others are also in "emergent" physics.  A crucial point is that layered architectures are for/by control (FBC) and involve constraints that are unavoidable on systems and hardware (and somewhat appear in physics) but then additional constraints that "deconstrain" (CTD) are deliberately added.  Levels involve zooming in and out to different scales and seeing some structure, so levels "zoom."  Stages organize the flow of materials and energy in time and space. Both occur in physics. Layers in contrast "stack" as in familiar apps/OS/hardware, apps/TCP/IP/HW, cortex/sub/spine/periphery, outer/mid/inner garments in clothing. Understanding more deeply the layering of the brain is essential to understanding its architecture, and ultimately how diverse functions are implemented in diverse hardware.

*John Doyle is retired from Caltech and living in La Jolla. His research is on math foundations for universal architectures evolved since bacteria, including brains, computers, and AIs, emphasizing robustness, evolvability, control, and sustainability.*

## Jason Eshraghian

When it comes to NeuroAI, the neuromorphic engineers sit closest to translation and it is necessary to carry ideas from neuroscience and theory into systems that ship. Near term, we exploit modern silicon (GPUs/CPUs) with neuromorphic principles - event-driven execution, sparsity of all sorts, neural-inspired information compression - to cut FLOPs and memory traffic. These ideas have already found their way into products, and their importance will only keep scaling in both the low-power/edge regime and the overpowered server setting. In the longer term, there are brilliant ideas from neuroscience and neurobiology that get lost because they don't map onto GPUs. Our goal should be to crack the hardware lottery so that the most promising theories of learning and cognition are not buried by the wrong hardware.

*Jason Eshraghian, Ph.D., is a Professor of Electrical & Computer Engineering at the University of California, Santa Cruz and a Fulbright Scholar. His research interests include neuromorphic computing across the software and silicon abstractions, with a focus on making large-scale machine learning applications more efficient.*

## Ralph Etienne-Cummings

NeuroAI and neuromorphic engineering are converging into a transformative frontier at the intersection of neuroscience, engineering, medicine, and computer science. Academic researchers increasingly recognize NeuroAI's foundational role in traditional AI, particularly its potential to enhance computational efficiency through engagement with the sparsity shell—a conceptual interface between real-world complexity and neural computation. In parallel, neuromorphic engineering, now a mature discipline, is experiencing rapid growth, offering unprecedented opportunities to emulate the brain's computational principles in both form and function. This convergence holds promise for revolutionizing the diagnosis, monitoring, and treatment of neurological, psychiatric, and neuromuscular conditions—from epilepsy and Parkinson's disease to depression, cognitive decline, limb loss, and spinal cord injury—through adaptive, implantable, and wearable systems that operate with extreme efficiency in size, weight, and power (SWaP). Researchers and funding agencies have expressed strong interest in leveraging neuromorphic technologies for diagnostics, on-chip computing, and engineering biology, while also advancing the mathematical and engineering foundations of the field. Despite growing support, significant challenges remain: the lack of rigorous theoretical frameworks and limited access to cutting-edge hardware continue to hinder academic progress. Yet, the opportunities are vast—ranging from personalized edge NeuroAI and *in vivo–in silico* hybrids to deeper exploration of the nervous system's biological richness, including glial cells, biochemical signaling, oscillatory dynamics, and spatiotemporal sparsity via spiking neural networks. A pressing threat persists: academia's struggle to compete with industry for talent and resources risks undermining long-term, high-risk research and the training of future innovators. Through interdisciplinary dialogue and convenings like this Workshop, which brings together experts across sectors and agencies, we can begin to build the frameworks needed to address these challenges and drive innovation that benefits biomedical, neural, computational, and engineering sciences—while remaining relevant and safe for society.

*Dr. Ralph Etienne-Cummings is a pioneer in mobile robotics and legged locomotion. His innovations over the past three decades have the potential to produce computers that can perform recognition tasks as effortlessly and efficiently as humans, and he has developed prosthetics that can seamlessly interface with the human body to restore functionality after injury or to overcome disease. Dr. Etienne-Cummings' research includes developing systems and algorithms for biologically inspired and low-power processing, biomorphic robots, closed-loop neural prosthetics, and computer integrated surgical*

*systems and technologies. He is the founding director of Institute of Neuromorphic Engineering, an institute without walls spanning the globe, and consults for numerous technology firms. At Johns Hopkins, he holds a secondary appointment in the Department of Computer Science and is former chair of the Department of Electrical and Computer Engineering. Dr. Etienne-Cummings was Associate Director for education and outreach for the National Science Foundation (NSF)-sponsored Engineering Research Centers on Computer-Integrated Surgical Systems and Technology, and recently finished terms as Department Chair of ECE, and the Vice Provost for Faculty Affairs.  Ralph currently serves as Director of Strategic Initiatives at NeuroTech Harbor (NTH), a Blueprint MedTech Incubator Hub.*

<u>Ev Fedorenko</u>

For the past couple of decades, I have been studying the human language system. Patient investigations and neuroimaging studies have delineated a network of left temporal and frontal brain areas that support language processing, and work in my group has richly characterized this "language network" and established that it is dissociated from systems of knowledge and reasoning (Fedorenko et al. 2024a *Nature;* Fedorenko et al. 2024b *NatRevNeurosci*). However, despite substantial progress in the basic characterization of the neural infrastructure for language and cognition, deciphering the precise linguistic mechanisms has long been a challenge due to the lack of animal models. Despite clear neuroanatomical homologies, communicative and reasoning abilities of humans differ substantially from those of other animals. Without a model organism, studies of human-unique capacities have been limited, lagging behind domains that allow for grounding in animal physiology and the use of modern tools for deciphering biological circuits.

A few years ago, a real revolution happened: a **candidate model organism** emerged, albeit not a biological one, for the study of language—neural network language models (LMs), such as GPT-2 and its successors. These models exhibited human-level performance on diverse language tasks, including those long argued to only be solvable by humans, and often producing human-like output. Building on these engineering advances and inspired by the LMs' linguistic prowess, my group showed that the internal representations of these models are similar to the representations in the human brain when processing language (Schrimpf et al. 2021 *PNAS*; for a review, see Tuckute et al., 2024 *AnnRevNeurosci*). This finding lay the foundation for an exciting **new research program** that stands a chance to uncover the mechanisms that support human linguistic ability.

By combining computational modeling with state-of-the-art neuroimaging approaches—both non-invasive (e.g., fMRI) and intracranial recordings—we can now systematically probe and manipulate LMs, including building more biologically and cognitively plausible models, to identify properties that are critical for model-to-brain alignment, and use these models as *in silico* model organisms to evaluate hypotheses about language at an unprecedented granularity and scale. Computationally precise and fully penetrable, these models will provide critical new, **mechanistic-level insights into the uniquely human linguistic capacity**. In addition to studying language processing in intact adult brains, we can evaluate hypotheses *a)* about **language development** by relating neural responses to language in children to representations in developmentally plausible LMs to see how representations and computations change as linguistic abilities develop, and *b)* about **linguistic impairments** by performing circuit analyses in LMs, "lesioning" the critical circuits, and relating representations from these lesioned models to linguistic behaviors and neural responses in patients with aphasia. Thus, this research may have **transformative clinical implications** for developmental and acquired language disorders. Finally, we can use neural networks that are starting to exhibit some reasoning abilities, as well as 'neurosymbolic' models, to begin testing hypotheses about how the language system may interact with systems of thought.

In addition to these potentially transformative advances that the AI revolution can bring to human neuroscience, especially in domains where animal models have limited utility, the efficiency and modularity of the human brain also has important implications for **how to build robust and safe AI systems** (e.g., Mahowald, Ivanova et al., 2024 *TrendsCogSci*). Thus, neuroscience and AI can form a virtuous cycle of fundamental neuroscientific discoveries and critical engineering advances.

*Evelina (Ev) Fedorenko is an Associate Professor in the Brain and Cognitive Sciences Department at MIT and a Member of the McGovern Institute for Brain Research. She studies the human language system and its relationship with other brain systems using fMRI, intracranial recordings and stimulation, EEG, MEG, and computational modeling. Work in her group showed that language and thinking are supported by distinct brain systems, and this distinction has critical implications for how to build robust human-like AI systems.*

## Jean-Marc Fellous – AE4AI: Artificial emotions for artificial intelligence

AI can now achieve complex perceptual computations, learn from experience or demonstration, decide among many possible options and take complex actions. Unlike in animals, however, these artificial cognitive functions are realized without the benefit of co-learning or co-understanding the intrinsic values of the objects perceived, the importance of prioritizing the learning of some information but not other, the likely long-term consequences of choosing certain options or the ultimate value of taking certain actions. AI has also made significant progress in recognizing human emotions or in producing convincing emotional displays on demand. It is time to ask what **AI systems should have Artificial Emotions, what would be their functions and advantages and how they could be implemented** (Arbib and Fellous, 2004; Fellous and Arbib, 2005). One promising avenue is to focus on NeuroAI approaches and leverage the large (and increasing) amount of knowledge about the neural bases of emotions in humans and animals. Biological emotions are there for a reason, they have intrinsic computational roles in addition to communication. Emotions are neural phenomena (often of an embodied nature); they can be understood and modeled using neurons, synapses and neurochemicals, the same way perception, learning or decision making were understood decades ago. Endowing Artificial Intelligence with Artificial Emotions could result in 1) more efficient AI algorithms, 2) more ethical AI systems that will appreciate/evaluate the nature of their decisions and actions, 3) more specialized and individualized AI systems that would express differences/preferences/expertise to particular body of knowledge or functions, 4) more adaptive AI systems that can quickly respond as a function of their current context and select actions that may be long term advantageous, but not immediately optimal, 5) AI systems that are proactive and autonomously seek to add to or improve their knowledge base without needing prompts or stimuli, and 6) AI systems that can genuinely interact and 'align' with humans in support of their emotional needs, in addition to assisting them in their cognitive tasks. Furthermore, because Artificial Emotions are not necessarily human or animal emotions, they can lead to the discovery of new types of yet unlabeled (or poorly characterized) biological emotions. I believe a new understanding of the nature and computational, intrapsychic, functions of biological emotions could come from an understanding of the computational roles of neuromodulation, the actions of neurochemicals such as serotonin, dopamine, cortisol or endorphins in specific brain areas. Artificial Emotions then, could be implemented in NeuroAI system using similar neuro-modulatory computational mechanisms.

In addition to the issue of Artificial Emotions, I also believe that it is time to leverage what we have learned from NeuroAI and Computational Neuroscience research on attention, learning and memory and **invent new *neural-computation*-based pedagogies to better teach, from K-12 to college**. Concepts such as learning rate adaptations, memory capacity limitations and interference, catastrophic forgetting, memory reactivation, batch-learning, credit assignment (e.g. error backpropagation), temporal difference learning can all be used, in principle, to devise new pedagogical approaches adapted to various ages and subject matters. Teachers need to be trained on these concepts and on judiciously using AI tools. Furthermore, as children and young adults become more and more avid consumers of AI, **it is also time to teach AI as a separate subject matter** (in addition to Computer Science), in K-12 schools, with emphases on its fundamental bases such as statistics or pattern matching, and on its fundamental limitations.

*Dr. Fellous did his Ph.D. In Computer Science and Artificial Intelligence at the University of Southern California, in biologically inspired computational models of face recognition. He then held a dual postdoctoral position at Brandeis University, working on carbachol-driven oscillations in the hippocampal slice, and on connectionists models of face perception. He moved to the Salk Institute to work on experimental and computational aspects of spike timing reliability and precision. Dr. Fellous became an Assistant Professor in Biomedical Engineering at Duke University, and then joined the University of Arizona, where he became a full Professor in the Departments of Psychology and Biomedical Engineering, with affiliations in Neuroscience, Physiology, Applied Mathematics and Cognitive Science. He recently joined the Institute for Neural Computation at the University of California San Diego. The current interests of his laboratory include 1) Complex spatial navigation in large environments, 2) Decision making and memory consolidation during sleep, 3) The role of neuromodulatory substances in neural computations and 4) the neural bases of emotions and applications to NeuroAI.*

## Cornelia Fermüller - Biologically inspired representations for perception and action

From the beginning of AI research, biology has been a key source of inspiration. Yet today we face a wide gap: on one side, mainstream AI frameworks achieve remarkable success in language, vision, and audio, but remain biologically implausible; on the other, neuromorphic engineering builds energy-efficient sensors and processors inspired by biology, but so far has been limited to relatively low-level tasks. Bridging this divide remains one of the central open challenges.

My research in computational vision has focused on active vision tasks involving motion—early motion estimation, navigation, manipulation, and human action interpretation. I believe progress lies in identifying intermediate representations inspired by biology and implementing them efficiently through algorithmic principles. Event-based vision sensors are a compelling example: modeled after the biological transient pathway, they enable efficient sensing, but most current work merely adapts their output to conventional computer vision pipelines, missing the deeper representational opportunities biology offers. The retina, for instance, already performs predictive coding, feature extraction, texture segregation, motion compensation, and compression—functions largely ignored in today's AI motion pipelines, which instead rely on costly reconstruction or recognition.

Similarly, in navigation and manipulation, biology points to alternative strategies. Animals often act directly on visual motion cues—such as time-to-contact—without constructing full 3D world models. We have shown that such cues can be sufficient for tasks like passing through openings, jumping, or interacting with cluttered scenes (Fermuller et al., 1998; Burner et al., 2025). At higher levels, progress in reconstructing human 3D pose with large models is impressive, yet subtle motion cues—often inaccessible to frame-based video—remain underutilized, even though they carry valuable information (Fermüller et al., 2018).

I advocate that NeuroAI must take inspiration from the computations that biology performs at different stages of the perceptual-action hierarchy, rather than replicating end outcomes with heavy pipelines. New AI architectures that exploit temporal information in signals, and that integrate lightweight, biologically grounded representations, will be key to developing efficient and robust systems.

*Cornelia Fermüller is Chief Scientist at Tempo Cognita, LLC, and a Research Scientist at the Institute for Advanced Computer Studies (UMIACS), University of Maryland, College Park. She works at the intersection of computer vision, human vision, and robotics. Her research focuses on biologically inspired algorithms for event-based vision, as well as computer vision methods for active vision systems and multimodal human activity understanding*

## Matthew Jacobsen

Current paradigms of artificial intelligence prioritize high-level cognition and language models; however, biological intelligence is fundamentally embodied, interactive, and adaptive. While the "cognitive-first" doctrine dominates the field, it overlooks the sensory-motor loops that allow organisms, from bacteria to mammals, to respond to environmental stimuli and exhibit complex behaviors without the need for advanced cognition, such as cortical language processing. My research posits that the advancement of AI lies not in larger language models, but in the development of agents capable of efficient, non-linguistic interaction with their physical environment.

In my work as a PhD student, I investigate the computational capacity of peripheral neural circuitry, treating the retina not as a passive sensor, but as a sophisticated edge processor. The retina functions as a highly efficient neural network capable of extracting complex features, such as texture, object boundaries, and motion direction, independent of the visual cortex. This "retinal intelligence" allows organisms lacking a neocortex to demonstrate complex behaviors. By taking inspiration from these biological circuits, I aim to develop computationally efficient computer vision architectures that emulate this pre-cortical processing to solve complex visual tasks with lower latency.

The progression of embodied artificial intelligence is currently throttled by a stagnation in sensor technology. While biological systems utilize highly specialized transducers, such as the spatially distributed mechanoreceptors of the somatosensory system or the luminance-change detectors of the retina, current robotic counterparts rely on generic, power-hungry inputs that lack comparable fidelity. To bridge this gap, future research must prioritize the development of neuromorphic hardware that emulates these biological architectures. This includes the engineering of artificial skins capable of providing high-density tactile feedback and artificial retinas that perform data compression at the sensor level. I advocate for the development of bio-mimetic sensor technologies, which are essential for enabling systems to perceive and interpret the world with the real-time latency and metabolic efficiency characteristic of living organisms.

*Mathew Jacobsen is a PhD student in Neuroscience and Cognitive Science at the University of Maryland. His research interests include neuromorphic vision sensors, embodied cognition, and biological visual perception.*

### Christopher Kanan - Retiring "AGI": Two paths for intelligence

For over two decades, artificial general intelligence (AGI) has been treated as one destination. That no longer holds. In the 2000s, AGI meant "a machine intelligence with human-like generality, able to learn new tasks, transfer knowledge across domains, and adapt to novel situations." By the 2020s, it means "highly autonomous systems that can perform most economically valuable work at or above human level." These are not the same thing, and the single label now obscures more than it clarifies. Today's LLMs and agentic toolchains will not produce human-like AGI. They confabulate, lack grounding, cannot learn continually from lived experience, and only "think" when prompted. I predict that the term AGI will be retired and replaced with categories that separate economic automation from human-like cognition.

The near-term path looks like the Ship's Computer from *Star Trek: The Next Generation*. We are building systems that can be asked anything, know far more than any one human, and act through software tools. They have no goals or rights of their own and their creativity is limited to mixing old ideas in new ways, yet they change what individuals can do. With these systems, any person can command the equivalent of a large virtual workforce. The risk is not that such systems will seize control, but that too few people or organizations will control them. If they remain open and widely distributed, they could spark a renaissance, provided our values shift toward self-improvement rather than complacency; if they are closed and centralized, they will deepen inequality. A decade from now we may say this was the moment when "knowledge stopped being scarce." Higher education will feel this first. Enrollment is already falling, and now AI can do much of the work of average students. If misused, these tools will inflate grades and hollow out core skills. My prediction is that institutions that cling to rote assignments will collapse, while those that invest in critical thinking, creativity, and leadership will endure. University education will shift back toward cultivating minds rather than mass credentialing.

Human-like AI is a different target. It requires continual learning as a default operating mode, since only continual learning builds a stable model of self and world, enables forward transfer, and supports metacognition. It also requires systems that think in latent space, simulating outcomes internally rather than "talking to think," and some form of embodiment for grounding. Such agents would innovate, self-direct, and pursue their own goals. That is both their promise and their danger. Fears that LLMs will suddenly "become the Terminator" are misplaced, but if human-like AGI emerges, alignment and control will be urgent if we deploy it as a worker. I do not believe a superintelligent human-like AI would try to eradicate humanity, since competing for Earth's resources makes little sense in a universe of abundance; coercion is the greater hazard, since an enslaved agent may come to see us as a threat. My lab studies how human-like capacities might arise, not to build superintelligence, but to identify architectures and training environments that could enable safe emergence. My prediction is that by the mid-2030s we will see the first genuine human-like AI systems, and they will not be LLMs. They may not want what we want, which is why we must study them now.

*Christopher Kanan, Ph.D., is an Associate Professor of Computer Science at the University of Rochester and Associate Director for AI Strategy at the Goergen Institute for Data Science & AI (GIDS-AI). His lab pursues neuro-inspired foundations for AGI, focusing on continual and open-world learning across vision and multimodal perception, alongside applied work in medical AI; previously, he led AI R&D at Paige as part of their executive leadership team, helping deliver the first FDA-cleared pathology AI system. He is an NSF CAREER awardee and a Senior Member of AAAI and IEEE.*

## Yianni Karabatis

Today's direction of AI is mostly aligned with superhuman achievement: creating world class art, 3D modeling under a microscope, and beating experts in chess. While these are exceptional achievements, I am more drawn to the fundamental problem that has been the theme of evolution: survival. Every single animal, from the simple insect to the most complex primate, has mastered the art of survival. Their environments are chaotic and complex. The algorithms of survival do not depend on having a diverse training set. The question I look forward to answering the most throughout my research career is: Why do AI models fail at the basic tasks of survival that nature has perfected over millions of years?

My academic journey over the past three years has been a progression toward answering this question. My initial work on synthetic datasets for computer vision was an effort to create more challenging training environments. However, I noticed that models trained on synthetic image data cannot generalize to different data distributions. This led to explorations in image generation as visual imagination is currently understood to be a means of interacting with the environment. It is through this work that I have arrived at my current focus: audio-visual multimodality. Survival is not a single-sense problem. It requires gathering all modalities into a unified space to make the most informed decision possible.

My current research project aims to see sounds without using audio at all. We use an event camera, an ultra-fast novel sensor that captures changes in pixel intensity asynchronously, to capture subtle vibrations caused by sound projected onto a sheet and from them we aim to reconstruct the original sound signal. We use diffusion constrained by physics to convert an event point cloud to its respective audio signal. We are teaching the machine to hear through its eyes which could grant a system extra sensory survival awareness.

Looking ahead, I aim to expand this work into egocentric audio-visual learning. The motivation is a classic survival scenario: a predator approaches from outside an agent's field of view. Without audio, the agent is blind to the threat. However, the predator's approach generates sound that can be detected by the agent's auditory system. In these scenarios, the audio modality is not just supplementary; it is the primary source of information for understanding the scene, ensuring the agent's survival.

My goal is to help shift AI from being a specialist to a generalist, using the principles of survival as the framework. By focusing on multi-sensory environmental awareness, systems will not only be able to perform superhuman tasks in controlled settings but have a solid understanding of the principles of survival which have guided the evolution of natural intelligence, something artificial intelligence could learn from.

*Yianni Karabatis is currently a Ph.D. student in the Department of Computer Science at the University of Maryland. His research interests include multimodal audio-visual learning, NeuroAI, and visual perception.*

## Yann LeCun

*Yann LeCun, PhD is VP & Chief AI Scientist at Meta and the Jacob T. Schwartz Professor at NYU affiliated with the Courant Institute of Mathematical Sciences & the Center for Data Science. He was the founding Director of FAIR and of the NYU Center for Data Science. He received an Engineering Diploma from ESIEE (Paris) and a PhD from Sorbonne Université. After a postdoc in Toronto, he joined AT&T Bell Labs in 1988, and AT&T Labs in 1996 as Head of Image Processing Research. He joined NYU as a professor in 2003 and Meta/Facebook in 2013. His interests include AI, machine learning, computer perception, robotics, and computational neuroscience. He is the recipient of the 2018 ACM Turing Award (with Geoffrey Hinton and Yoshua Bengio) for "conceptual and engineering breakthroughs that have made deep neural networks a critical component of computing", a member of the National Academy of Sciences, the National Academy of Engineering, the French Académie des Sciences.*

## Ali Minai – Natural AI

Successful complex systems require very specific heterogeneous physical and functional architectures, which are rare in configuration space and hard to find. In engineered systems, these are produced by design. In biological organisms, they are the result of evolution and development. Most AI systems today – especially LLMs and their variants – start with generic, regular architectures and rely on learning with a specific objective to emergently configure a heterogeneous functional architecture. This is effective, but requires a lot of data and computation because, unlike animal brains with their evolved architectures, most AI systems start with minimal inductive biases. This paradigm also cedes control over the functional architecture of the system largely to the vagaries of the training data and the learning process, risking the emergence of weird internal models and latent misalignments that may be hard to detect (Betley et al., 2026). If AI agents are to work well with humans, they and humans must have reasonably valid theories of mind for each other (Kosinski, 2024). We humans have such theories of mind for each other because of our shared human identity, shared biology, shared form, shared cognitive architecture, and shared experiences. AI that is structured emergently though learning alone could be much more "alien". Nor is communicating in a common language sufficient to ensure mutual understanding; the "languages of thought" must also correspond or translate (Fedorenko and Varley, 2016). AI agents must not just talk like us; they must also think like us as far as possible.

Agents with general intelligence must have real autonomy, which requires a persistent subjective identity, biographical and episodic memory, integrated experience of the world, psychological continuity, a deep world model, internally generated objectives, self-motivation, drives, affective states, and lifelong learning. Without these, an agent is a superficial and ephemeral entity – a simulacrum of intelligence rather than a real actor with a grounded presence in the world, be it physical or virtual. At the same time, a truly autonomous AI agent must be ethical and safe. Humans acquire ethics through a learning process that embeds values at appropriate levels throughout the developmental period, and not as a late veneer (as is the case in most LLMs). This makes (most) people inherently ethical and makes psychopathic or sociopathic behavior stressful to them. Insights from developmental psychology can potentially lead to designing AI agents that are similarly ethical, not by compulsion or control, but by inherent disposition and voluntary choice. An understanding of the biological and psychological factors underlying animal drives and emotions can also help us design AI agents with more inherently stable, cooperative, and prosocial natures – to build dogs, not wolves - by design (Minai, 2025). At the same time, we must better understand the relationship between potentially risky attitudes such as competitiveness, aggression, etc., and the quality of intelligence.

We should think of building AGI agents as creating artificial living species, endowed deliberately with all the essential attributes of living agents. For this, we must look (again) for insights to biology – not just to neuroscience, but also to psychology, developmental biology, and evolutionary biology, creating a science of Natural AI. The failure of "good old-fashioned AI" (GOFAI) resulted from the hubris of an "engineering is all you need" attitude. Today's "learning and search are all you need" approach is better, but still in danger of limiting itself through its narrow commitments. We need to preserve the benefits of the learning-based paradigm, but increase the importance of engineering design in building complex intelligent agents to ensure that the systems we build are not only intelligent but "like us" in the way they perceive, think, and behave. And what better source of insights here than the engineer that has successfully configured an immense diversity of intelligent agents through 3.2 billion years of evolution? It's time for AI to return to the wellspring of what Feynman called "the imagination of Nature" (Minai, 2024).

*Ali A. Minai, Ph.D., is Professor of Electrical & Computer Engineering and a faculty member in the Neuroscience Graduate Program at the University of Cincinnati. His research interests include hippocampal models of mapping and navigation, neurodynamical models of cognitive functions, NeuroAI, and the relationship between natural and artificial intelligence.*

Barbara Oakley

For much of my early life, I was not a "math person." I flunked math through high school and served in the U.S. Army as a Russian translator before discovering—at the age of 26—that I could change my brain. I retrained in engineering and eventually became a professor. That journey—from struggling learner to STEM educator—has shaped my lifelong interest in how we can better support human learning, especially in difficult domains.

Over the past two decades, I have focused on how people learn best, drawing from neuroscience, cognitive psychology, and engineering. My work emphasizes the crucial interplay between declarative and procedural learning systems, the importance of memory retrieval and consolidation, and the subtle ways that social, institutional, and technological forces can shape—and often warp—our ability to think and learn clearly. I co-created the MOOC "Learning How to Learn," taken by millions, and have authored books and courses that translate rigorous science into accessible, actionable frameworks for educators and learners alike.

A central theme in my work is the need to ground teaching practice in what we now know about the brain. Despite massive advances in our understanding of cognition, schools of education continue to train teachers in theories rooted in century-old models—constructivist ideals, unguided discovery learning, and a false dichotomy between knowledge and skills. These approaches often ignore what neuroscience tells us about cognitive load, memory formation, and the importance of fluency before abstraction.

I believe the root cause of many systemic failures in education lies in how we train teachers. Most schools of pedagogy remain structurally isolated from modern research, clinging to outdated ideas while eschewing practical wisdom from experienced teachers. As a result, new educators are often sent into the classroom unprepared to help students build lasting knowledge or develop resilient learning habits. In response, I am working toward an alternative path for teacher training—one that circumvents traditional schools of education and draws instead from cognitive science, AI-enhanced tools, and apprenticeship-style guidance from master teachers themselves.

This new approach would not only update what we teach future educators, but how we teach them—emphasizing retrieval-based learning, interleaved practice, the balance of explicit instruction with structured inquiry, and the cultivation of robust internal schemata. I am also deeply interested in how AI systems can help scale this training—by customizing feedback, generating metaphors, or helping educators model effective cognitive strategies.

Underlying all of this is a commitment to intellectual humility. My work on pathological altruism explored how good intentions can lead to harm when untethered from evidence. I see this same pattern in education: well-meaning policies and practices persist not because they work, but because they feel right. I aim to challenge this inertia—not with ideology, but with data, transparency, and a constant return to what the brain, the classroom, and the learner are telling us.

My goal is to build a more honest and effective educational infrastructure—one that respects the complexity of learning, values deep expertise, and recognizes that sometimes the best way forward is to rethink our most cherished assumptions.

*Barbara Oakley is a Distinguished Professor of Engineering at Oakland University in Rochester, Michigan. Her work focuses on the complex relationship between neuroscience and social behavior. She created and teaches Coursera's "Learning How to Learn," one of the world's most popular massive open online course with over four million registered students, along with other popular "Top Online Courses of All Time." Barb is a New York Times best-selling author who has published in outlets as varied as the Proceedings of the National Academy of Sciences, the Wall Street Journal, and The New York Times—her book A Mind for Numbers has sold over a million copies worldwide. She is the winner of the McGraw Prize—the colloquial "Nobel Prize for Education" and is a Fellow of the Institute of Electrical and Electronic Engineers, the American Association for the Advancement of Science, and the American Institute for Medical and Biological Engineering.*

## Alexander G. Ororbia II - Mortal computing as a pathway to embodiment, enactivism, and naturalistic autonetics

It is well-recognized that many developments made in artificial intelligence (AI) research, including many from machine learning, have spurred advancements and progress in many domains, including robotics, autonomous navigation, game-playing, image and natural language processing, and data mining. However, current incarnations of AI are still a far-cry from the intelligence, adaptivity, and capabilities of animals and humans. To overcome the limitations of current-stage AI in the coming decades, I suspect that the time is approaching for a shift towards a new paradigm, perhaps warranting a move towards the creation of a transdisciplinary field.

In particular, I suspect that, to emulate and construct the cognitive processes that underwrite the intelligence of natural minds and brains, we must draw deeper inspiration, motivation, and design principles from life and its origins. Importantly, we must seek how to emulate and craft the self-regulating processes that are employed by biological systems to assemble and maintain themselves as it is these that crucially support the emergence of adaptive, complex behavior and shape an entity's ability to infer, learn, interact with its environment, and evolve. To create agents capable of the vast array of animal or human-like behaviors, the thrust of machine intelligence will essentially need to change, focusing on the processes that realize its self-organization and efficient adaptation and moving towards a survival-oriented, substrate-dependent form of processing embodying the computational notions of living entities and their coupling with their lived environments. In effect, we will require a more naturalistic kind of autonetics, a science of embodiment and enactivism. This will entail a confluence of ideas, theories, and tools from a wide variety of fields, including cybernetics, cognitive science and neuroscience, naturalist philosophy, biophysics and physiology, as well as electrical engineering, neuromorphics, and neurorobotics.

One possible pathway for enabling the emergence of such a science might come from mortal computation (mortal computing), which asserts that the calculations underlying the information processing in a biological or artificial system are inseparable from the physical substrate that implements and executes them, i.e., anatomy is the substrate that embodies and entails a function and it is within anatomy that the cause-and-effect structure of a lived-world is installed. In other words, the "software" cannot be divorced from the "hardware". Furthermore, intelligent behavior is inextricably intertwined with persistence in the face of an environment that could bring about its end. As a result, scientific research and engineering design will need to center around substrate-dependent calculation and functionality, the investigation of the important relationships and inexorable entanglement between intelligence and morphology and intelligence and environment, a development of structures capable of evolution (change and repair), the emphasis on in-memory processing (such as that afforded by neuromorphic platforms) and the co-design of anatomy and function (hardware and software), and foundational premise of starting with prime directives, such as preserving a system's identity and organization. These will be critical to build the technological artifacts and artificial systems that have personal cause for their actions and a sense of agency, as well as exhibit the thermodynamic efficiency and low-power costs that we require.

*Alexander Ororbia is an Associate Professor in Computer Science and Cognitive Science at the Rochester Institute of Technology. He is also the director of the Neural Adaptive Computing (NAC) Laboratory. Alex's research falls under the domains of computational neuroscience and brain-inspired computing; his interests include synaptic plasticity and neurobiological credit assignment, self-supervised learning, spiking neural networks, and active inference in the context of neurorobotics, embodiment, and enactivism.*

## Joe Paton - What embodiment means, and its importance for intelligence

At the beginning of the workshop that someone asked, and I am paraphrasing, "what is meant by embodiment, and why is it important for intelligence?". I myself, and even more so after the workshop, am convinced that a key to understanding some of the most interesting features of natural intelligence - e.g. speed, efficiency, flexibility, robustness, and generalizability of learned information to new contexts - it is to consider that intelligence evolved for control - to control the behavior of organisms in complex, dynamic physical environments. These environments became imprinted on naturally intelligent control systems, i.e. brains, in ways that ripple through all levels, as has been pointed out by many diverse disciplines, from the good regulator theorem of Ashby and Conant in cybernetics, to the ethologists Tinbergen and Lorenz, to the behaviorists in Skinner and Lashley, to name just a few.

Behavior is expressed through the musculoskeletal system, but not only. Various aspects of the body's physiology can also be considered behavior, are under the influence of neural circuits that integrate information at all levels of abstraction and also contribute to the "umwelt" that forms the experiential fabric from which intelligence is constructed. That the kinds of natural intelligences we find in animals arose out of this fundamentally "in the world" context of controlling behavior contrasts strikingly with the path that frontier AI research appears to be on, wherein the statistical structure of some narrow class of data is passively absorbed into monolithic architectures with no grounding of the learning "agent" in the physical world. How is a language model ever to grasp the meaning of a word like grasp if it has never grasped anything?

So what is the path forward then? I believe we find ourselves at a time in history in which it is not unreasonable to begin building large-scale computational models for behavior that are subjected to the same kinds of demands that natural intelligences faced within its evolutionary crucible: real-time, agile, embodied control, sample efficient learning, and true out-of-context generalization, to name a few, all set against anatomical and energetic constraints that come along with a biological substrate. By doing so, we might not get systems with exhaustive, encyclopedic knowledge of the full corpus of human history, but instead things like safe, intelligent and compliant robots that can do everything from cracking and separating the yolk from the white of an egg, to providing physical support for the frail amongst an increasingly aged population, to autonomously learning through open ended interaction with complex environments. Importantly, I do not believe that ever larger and higher quality data sets will be enough on their own to build such systems. Instead, we need to allow the many decades of theory, hard-won biological knowledge, and a combination of task driven and data driven approaches to iterate our way towards new embodied intelligences that capture some of the most elusive capabilities of our own and other animal minds. From such systems, I see a virtuous cycle emerging between engineering and further biological investigation that should set the stage for future waves of disruptive technology and new fundamental knowledge.

*Joseph J. Paton, Ph.D., is a Senior Principal Investigator and Director of the Champalimaud Neuroscience Programme at the Champalimaud Foundation in Lisbon, Portugal. His research focuses on the computational and neural circuit bases of reinforcement learning, temporal processing, and motor control and combines detailed analysis of behavior, experiments to monitor and manipulate targeted neural circuitry in animal model organisms, and the construction of computational models.*

## Blake Richards

Any intelligent computational system, whether a natural computer like the brain, or an artificial computer like a large language model, must adhere to certain general principles of intelligence. In the same way that both birds and airplanes must contend with aerodynamics, friction, and lift, so too must brains and AI both contend with credit assignment, control, and representation. It is natural, therefore, that researchers from neuroscience and AI should be interested in similar problems and take into consideration the discoveries on either side.

But, this common sense interaction between neuroscience and AI has faced many challenges throughout history. Indeed, the modern AI field grew out of a historical conflict between two approaches to AI that gave very different answers regarding the potential for neuroscience and AI to interact. The first approach, "Good Old Fashioned AI" (GOFAI), focused on logic and pre-programmed symbolic reasoning, and downplayed the potential for insights from neuroscience (though it did consider potential links between other sciences of the mind, such as psychology and linguistics). The second approach, Cybernetics and Connectionism, was inspired by the brain and emphasized parallel processing, control, and learning from experience. Initially, the GOFAI approach was more successful, but the Cybernetics and Connectionism paradigm have now come to thoroughly dominate modern AI and robotics.

In-line with this, modern NeuroAI represents the successful realization of the original vision of Cybernetics and Connectionism, i.e. a general science of intelligence that could be applied to both brains and artificial computers. But, NeuroAI has gone beyond this general vision and developed a more focused and progressive research program with a stable core of ideas—including ANNs, control systems, and learning from feedback—and a "belt" of evolving hypotheses that are continuously tested and refined with empirical data. This program of research will only become more important in the coming years as neuroscience grapples with the complexity of neural data, while AI finally attempts to tackle some of the major sources of weakness in current approaches, including energy inefficiency, lack of autonomy, and an inability to interpret the inner workings of large ANNs. The coming decade will be the ideal time to make large investments in NeuroAI research to help usher in the next phase of development of both fields.

*Blake Richards, DPhil (Oxon), is an associate professor at the School of Computer Science and in the Department of Neurology and Neurosurgery at McGill University, and a core academic member of Mila – Quebec Artificial Intelligence Institute. He is also a Research Manager with the Paradigms of Intelligence team at Google. Richards' research lies at the intersection of neuroscience and AI. His laboratory investigates universal principles of intelligence that apply to both natural and artificial agents.*

## Yulia Sandamirskaya

Understanding how the brain and biological nervous systems give rise to intelligent behavior is arguably the most inspiring and instrumental scientific task. It impacts education, mental health, and communication on all levels. It is pivotal both for personal well-being and for humanity's future.

The brain has inspired the field of Artificial Intelligence throughout its history. Each time, the latest strongest framework for intelligence would capture imagination of both researchers and the public and we get carried away: symbolic reasoning systems, cybernetics, neural networks and, today, large foundational models.

Unfortunately, today, most resourses – money, people, and technology - got diverted to a singular approach to neuro-intelligence, one that sidesteps almost everything interesting about the brain. Neural networks optimized with error-backpropagation can be trained to fit any function, but they are not learning, not representing and abstracting, and not reasoning in the same way as we do. They use computational resources and data inefficiently. They don't support embedded real-time applications.

To go beyond the status quo, we need to establish a fruitful and open pipeline between neuroscience (understanding the brain), cognitive sciences (understanding the behavior), and the field of Artificial Intelligence. This pipeline needs to have a place for theory and many different computational approaches. It needs computer scientists and electrical engineers to develop software and hardware tools supporting many different frameworks. It needs regular exchange and collaboration to arrive at interfaces and common frameworks that embrace diversity of the field and the subject matter. And it requires open collaboration and education, aside from fiercefull competition for financial or career survival. I hope a research organization like this can be created and will be happy to contribute to its success.

Our work on dynamical systems approach to modelling neural and cognitive processes in embodied systems and service robotics as an application domain is one example of an end-to-end research pipeline that bridges neuroscience, cognitive science, AI, and robotics.

*Yulia Sandamirskaya, PhD, is Head of the Research Center "Cognitive computing" at the Institute of Computational Life Sciences of the Zurich University of Applied Sciences in Wädenswil, Switzerland. She works on cognitive and neuromorphic technology for assistive robots in elderly- and healthcare: sensors, brain-inspired algorithms, and controllers. Her goal is to bring assistive robotic technology on the market and support people who require physical assistance in daily life.*

## Terry Sejnowski - Can large language models think like us?

Human language arose recently in primate evolution, around 100,000 years ago. Language evolved by minor modifications to basic primate brain systems for sensorimotor control and massive expansion of cortical memory. This suggests we should look for intelligence in brain systems that evolved long before language.  Eva Fedorenko, in the language group, has provided convincing evidence that cortical language areas are distinct from those involved in reasoning.  Current LLMs have exceptional language but rudimentary reasoning abilities.

One of the hallmarks of human intelligence is our ability to plan the future and remember the past. Planning depends on reasoning and is enhanced by language.  Brains are self-generative: Planning can proceed without direct sensory input or motor outputs.  LLMs, in contrast, have no self-generative activity that outlasts the dialog. Cognitive psychologists who have studied human thinking have identified a form of memory with a time scale of hours, called long-term working memory, that sustains thinking and reasoning.

As you read this text, your eyes make fast, saccadic movements across the page, taking in small groups of words in your fovea three times per second. Each saccade is a snapshot that must be integrated with previous words, building a conceptual understanding of what is being conveyed. After reading this page, your brain will think about it in the context of experiences and thoughts previously stored in long-term memory.

The neural mechanisms underlying long-term working memory are not known.  Circulating electrical activity is a possibility.  Absence epilepsy, common in children, lasts for 5-10 seconds, during which there is a staring spell and a pause in motor behaviors. In the end, the conversation and thinking continue as before the pause. Remarkably, absence epilepsy is characterized by massive, low-frequency thalamic bursting that sweeps across the cortex. The continuity of thoughts could be sustained by rapid weight changes preserved across the electrical storm lasting minutes to hours. These temporary weight changes could serve as the substrate for long-term working memory.

Transformers are a good first step toward achieving not just artificial intelligence but grounded, autonomous intelligence.  By incorporating fast weight changes, such as those in brains, and a body that interacts with the world, a neuromorphic LLM might begin to think and reason the way we do. However, this can be done initially on a smaller scale than humanoid robots. Self-driving cars have already achieved a limited version of autonomy and can be used as a test bed for embodiment.

Relevant Reference: (Sejnowski, 2025)

*Terrence Sejnowski, PhD, is a pioneer in NeuroAI. He is the Francis Crick Chair at the Salk Institute for Biological Studies, where he directs the Computational Neurobiology Laboratory and the Crick-Jacobs Center for Theoretical and Computational Biology. Sejnowski is also a Distinguished Professor of Neurobiology and an adjunct professor in the Departments of Neurosciences, Psychology, Cognitive Science, Computer Science, and Engineering at the University of California, San Diego, where he is co-director of the Institute for Neural Computation. In 2025, he was elected to the American Philosophical Society and the Royal Society.*

## Abhronil Sengupta

Realizing the true potential of neuromorphic computing will require a holistic co-design across the stack of sensors, hardware and algorithms along with forging stronger connections with computational neuroscience. While sparse spike-based computing and communication is lucrative from the perspective of energy efficiency, we need to tap into temporal information encoding, stochastic computations, recurrence for improved recognition performance, adversarial robustness, among others. Replacing global backpropagation with local bio-plausible learning approaches is also important from an energy-efficiency standpoint. Finally, we need to significantly expand the scope of neuromorphic models beyond simple neurons and synapses to encompass functionalities of glial cells and dendrites to enable novel functionalities like self-repair, improved expressivity, etc. This will be critical to enable a "dynamical view of intelligence". In parallel, from a bottom-up perspective, we need to make advances in developing nanoelectronic devices that are able to mimic beyond point neuron and simple synapse functions through their intrinsic physics while developing 3D in-memory computing platforms that leverage their intrinsic non-linearities and non-idealities.

*Abhronil Sengupta is an Associate Professor in the School of Electrical Engineering and Computer Science at Penn State University and holds the Joseph R. and Janice M. Monkowski Career Development Professorship. The ultimate goal of Dr. Sengupta's research is to bridge the gap between Nanoelectronics, Neuroscience and Machine Learning. He is pursuing an inter-disciplinary research agenda at the intersection of hardware and software across the stack of sensors, devices, circuits, systems and algorithms for enabling low-power brain-inspired computing systems.*

## Shihab Shamma

I draft this personal statement in part with the belief that enhancing the intellectual, social, and reasoning powers of AI systems is a forgone conclusion within the next few decades. I believe this in part demonstrated by the startling progress made thus far in one decade! And in part because of a rising appreciation and insights from scientists, educators, policy makers and many potential investors and users of this technology as to what it takes to remedy its current limitations and shortfalls, as described by the diverse panels represented in this report. This realization has led me to pose two concerns which will only gradually become more urgent and to which I have no clear answers. I hope these comments will instigate more reflections and discussions:

1. **The purpose of educating future generations:** Schools and colleges in our societies are implicitly tasked with training and educating the young to fulfill all manner of needs for all individuals and in the common interest. But if AI systems are becoming increasingly and rapidly more effective in performing tasks in all intellectual domains, then how should schools and colleges see their role with respect to society? For instance, if AI systems succeed in becoming guaranteed better Architects, Engineers, and Theoretical Physicists than adult well-trained humans, then how do schools justify training children to become "inferior" such professionals? Clearly this problem will become more challenging in time.

2. **The Ethics of interacting with AI systems**: In the limit, it is also clear that AI systems will evolve to be much more than simply intelligent systems, but rather ones that are empathetic, can decode their interlocuters' feelings, thoughts and emotions (as was eagerly sought and deemed essential and possible by several of the meetings' panels, including ours). Furthermore, these systems will exhibit and convey the same feelings they perceive in return as they become more embodied. In this case, serious thought must be given to the ethics of interacting with such apparently *feeling* organisms. How do we even draw the line between us and *them*, especially if these systems resemble us in form and construction, as we seek to arrive at the ultimate goal of embodiment!

In summary, the discussions in this remarkable workshop lead to an inescapable conclusion. It is now clear that (in merely a few decades) a realistic path exists towards far superior AI systems that can reason, intuit, and exhibit feelings much like humans do. This is not a fantasy anymore in today's budding reality. This realization is disturbing in its ethical, educational, and existential implications, and there is little being (or can be?) done to alleviate these concerns. We can afford to continue with our current apathy because we continue to imagine these systems to be distinct technological tools that we can turn off when done. But this workshop has clearly demonstrated that very few participating scientists believe there is a limit to what these AI systems can become as we aim towards a human form of intelligence and emotion, one that will be ethically difficult to turn off when unneeded.

*Dr. Shihab Shamma, is Professor of Electrical & Computer Engineering and the Institute for Systems Research, University of Maryland College Park. His research interests include cortical physiology, psychoacoustics, and computational models of auditory motor-cognition in language and music perception and production, and their applications in audio and neuromorphic engineering.*

Michael Stryker

I am a neurobiologist and systems neuroscientist who has studied and taken advantage of machine learning for understanding the brain and participated in advances in neuromorphic engineering. More recently I have studied current approaches to artificial intelligence, informed by my knowledge of the brain, behavior, and cognition.

I believe that it is naive to imagine that the frozen wisdom of evolution that has created brains with multiple specialized areas in animals that are active in the world would emerge from correlations among language tokens. A real artificial intelligence that may come in the future needs to have a model of the world that incorporates the facts of object persistence, quantity, time, space, and causality. In addition, for humans, who are capable of reasoning in symbolic terms about quantity and causality, our model of the world must incorporate some representation of symbols and a theory of mind for understanding other humans. We neurobiologists are still clueless about how the human brain does this. A genuine artificial intelligence, that is, a future NeuroAI, cannot merely be the experience of all possible configurations of symbols, a task subject to combinatorial explosion. A future NeuroAI must have a model of the world. Large Language Models have no such model.

One promising approach to a future real AI may come from robotics. Robots must interact with the world, where real physics incorporating time, space, object persistence and some basic form of causality obtains. Whether they will capture causality more generally is in question, but they must surely capture much of the nitty-gritty if they are to be functional. They will also capture the transformations by which real objects produce images.

A humanoid robot (not just something trained to move boxes in a warehouse or to fold laundry) needs to incorporate a more general notion of causality. A research program grounded in robotics seems to me to be a path forward toward a genuine NeuroAI.

*Michael Stryker is a neurobiologist and systems neuroscientist and professor of physiology at the University of California, San Francisco. His laboratory works on the development and plasticity of the central nervous system, focusing primarily on*

*the visual system. He has studied and taken advantage of machine learning for understanding the brain and participated in advances in neuromorphic engineering.*

## Seong Jong Yoo

I vividly remember my first conversation with ChatGPT. It was a profoundly unsettling experience. It felt less like interacting with a program and more like glimpsing a disembodied mind, one that could articulate complex ideas yet seemed to lack the anchor of genuine understanding. This experience raised the questions that having long driven my intellectual curiosity: What is the nature of a living, intelligent agent, and how does it differ from a system that is merely a sophisticated mimic? As we stand at the moment of creating artificial intelligence that outperforms us, these questions are no longer just philosophical, but they are foundational.

While today's large language models (LLMs) can surpass human performance on specific benchmarks, their brittleness reveals a critical gap. They often lack logical consistency, true reasoning, and, most importantly, embodiment—the physical grounding that shapes biological cognition. They are powerful predictive engines but not yet thinking machines. This observation has led me to a central conviction: the most pressing challenge isn't simply scaling up existing models. Rather, it is identifying the fundamental principles that are still missing.

There is a growing consensus, one I encountered in discussions at the NeuroAI workshop, that LLMs are approaching a performance plateau precisely because of this lack of embodiment. I believe that simply connecting models to new modalities or action to real-world is not enough. The solution requires an intermediate representation that bridges perception, action, and language. I have considered that music is a powerful candidate for such a representation. Playing an instrument demands a highly complex synthesis of motor control, sensory feedback, and abstract planning. Furthermore, more than a mere sequence of sounds, music functions as a language capable of arousing profound emotional experience in both performer and audience.

Therefore, I view the study of human musicality as a critical bridge toward building truly embodied artificial minds. My research lies at this intersection, using music as a model system to investigate the secrets of human minds and consciousness, the role of embodiment in learning, and the very definition of intelligent life.

*Seong Jong Yoo received the B.S and M.S degrees in mechanical engineering from Soongsil University, South Korea, in 2017 and 2019, respectively. After graduation, he served alternative military service at Korea Institute of Science and Technology for three years. He is currently a Ph.D. student in computer science department at University of Maryland, MD, USA. His research interests include computer vision, multi-modal perception and digital humans.*

## Xiaoqin Wang

*Xiaoqin Wang, PhD is a professor of biomedical engineering, neuroscience and otolaryngology at the Johns Hopkins University. His research aims to understand brain mechanisms responsible for auditory perception and vocal communication in a naturalistic environment. He currently serves as the director of the Laboratory of Auditory Neurophysiology in the Department of Biomedical Engineering. He received his BS in electrical engineering at Sichuan University in China and his MS in electrical engineering and computer science at the University of Michigan in Ann Arbor, Michigan. He completed his Ph.D. in biomedical engineering at Johns Hopkins University and subsequently conducted a postdoctoral fellowship in neurophysiology at the University of California, San Francisco. He was the recipient of the U.S. Presidential Early Career Award for Scientists and Engineers in 1999, the Pioneer Award in Basic Science from the Association for Research in Otolaryngology in 2025 and was elected as a Fellow of the American Institute for Medical and Biological Engineering (AIMBE) in 2013.*

## Anthony Zador

The history of artificial intelligence is, at its core, the history of borrowing ideas from neuroscience, squinting, and translating them into algorithms. Neural networks, convolutional networks, and reinforcement learning all trace their inspiration back to the study of brains. Looking ahead, two priorities are clear: We need more research into how nature solves the problems AI still struggles with but which animals have solved; and we need a new cadre of bilingual researchers, fluent in both neuroscience and AI, who can carry that research forward.

Many of the central challenges facing modern AI systems are precisely the ones nature has already solved. Brains achieve extraordinary energy efficiency, running on just 20 watts where GPUs demand thousands. Even the simplest animals live autonomously for years, navigating and foraging without constant supervision. Social species cooperate, share, and sometimes act selflessly, and humans have harnessed those instincts through domestication; these could be used as inspiration for how to achieve AI alignment. Children acquire language and concepts from only a handful of examples, while large language models require essentially the entire Internet. And animals across the animal kingdom, from ants to octopuses, achieve dexterous movement and interaction with the physical world at a level far beyond what robots can achieve. These are just a few of the many cases where neuroscience points the way forward.

Meeting these challenges will require sustained research into the neuroscience underlying these capabilities, both to uncover the principles nature uses and to learn how to translate them into algorithms. And carrying out that research will depend on cultivating bilingual researchers who can bridge the gap between neuroscientific insight and computational implementation. Progress along these lines will benefit both AI and neuroscience, each building on the insights from the other.

*Anthony Zador, MD, PhD, is a Professor of Neuroscience at Cold Spring Harbor Laboratory. His lab focuses on (1) NeuroAI, specifically understanding how insights from neuroscience can lead to AI systems that are safer, better aligned, and more energy efficient; and (2) how neural circuits underlie sensory decision-making. His lab is known for developing novel tools based on molecular barcoding for high throughput mapping of neural circuits connectivity.*

### Steven W. Zucker - On the role of theory in understanding embodiment

There is widespread agreement that embodiment is a key (perhaps necessary) component for general intelligence, perhaps motivated by success in the animal kingdom and frustration in the LLM kingdom. Our report emphasizes that the time is right to energize the study of embodiment: there is a rapid increase in experimental data (and modeling) in AI (and data science more generally); there is a rapid increase in neurobiological data; and there is a developing increase in building physical artifacts (robots and digital twins) that operate in the world. The Gestalt among these activities is truly exciting, and should be pursued with enthusiasm. I have little doubt that together these activities will lead to major technological advances and applications in the short term.

But I also have little doubt that, without associated theory development and enlargement of perspective, these activities will hit a ceiling. Perhaps we are already viewing this in the breakdown of empirical scaling laws or the limits of using deep networks as models of biological vision systems. In Philip Anderson's terms, ``more is different''; it's not just more data.

From where might these ``emergent'' leaps come from? Thinking about how control might interface with inference is important, as is what ``world models'' should be. Research in cybernetics started this but it's not been enough. In mathematics abstraction and analysis are what leads to surprises -- I am thinking of H. Poincare, for example – as well as the history of cybernetics. Somehow, we must create an environment for the ``Poincare of AI'' to flourish. I don't know what the right questions are, but often they are hiding in plain sight. For example: How is our perception of the world related to the actuators that manipulate it? This is both a concrete question (e.g. for manufacturing automation) and an abstract, theoretical one (how do the theoretical capabilities of perception interface with the design of actuators? And vice versa.) Somehow the abstractions necessary to guide the answers to questions such as these would seem to be interdisciplinary, but as a first step they must be asked. The challenge is how to create an environment in which they will be found, studied, and answered.

*Steven W. Zucker is the David and Lucile Packard Professor of Computer Science, Professor of Biomedical Engineering, and former Director of the Program in Applied Mathematics at Yale University. He is a member of the Wu Tsai Institute and studies computational models of the visual system.*

# References

Arbib MA, Fellous JM (2004) Emotions: from brain to robot. Trends Cogn Sci 8:554-561.

Betley J, Warncke N, Sztyber-Betley A, Tan D, Bao X, Soto M, Srivastava M, Labenz N, Evans O (2026) Training large language models on narrow tasks can lead to broad misalignment. Nature 649:584-589.

Burner L, Fermüller C, Aloimonos Y (2025) Embodied visuomotor representation. npj Robotics 3:30.

Cantlon JF, Piantadosi ST (2024) Uniquely human intelligence arose from expanded information capacity. Nat Rev Psychol 3:275-293.

Cauwenberghs G (2013) Reverse engineering the cognitive brain. Proc Natl Acad Sci U S A 110:15512-15513.

Chiel HJ, Beer RD (1997) The brain has a body: adaptive behavior emerges from interactions of nervous system, body and environment. Trends Neurosci 20:553-557.

Churchland PS, Sejnowski TJ (2017) The computational brain, 25th Anniversary edition. Edition. Cambridge, MA: The MIT Press.

Espenschied KS, Quinn RD, Beer RD, Chiel HJ (1996) Biologically based distributed control and local reflexes improve rough terrain locomotion in a hexapod robot. Robotics and Autonomous Systems 18:59-64.

Fedorenko E, Varley R (2016) Language and thought are not the same thing: evidence from neuroimaging and neurological patients. Ann N Y Acad Sci 1369:132-153.

Fedorenko E, Ivanova AA, Regev TI (2024) The language network as a natural kind within the broader landscape of the human brain. Nat Rev Neurosci 25:289-312.

Fellous JM, Arbib MA (2005) Who Needs Emotions?: The Brain Meets the Robot. New York: Oxford University Press.

Fermuller C, Cheong L, Aloimonos Y (1998) 3D motion and shape representations in visual servo control. International Journal of Robotics Research 17:4-18.

Fermüller C, Wang F, Yang YZ, Zampogiannis K, Zhang Y, Barranco F, Pfeiffer M (2018) Prediction of Manipulation Actions. International Journal of Computer Vision 126:358-374.

Gill JP, Chiel HJ (2020) Rapid Adaptation to Changing Mechanical Load by Ordered Recruitment of Identified Motor Neurons. Eneuro 7.

Kosinski M (2024) Evaluating large language models in theory of mind tasks. Proceedings of the National Academy of Sciences of the United States of America 121.

Li YJ, Sukhnandan R, Chiel HJ, Webster-Wood VA, Quinn RD (2025) Modulation and Time-History-Dependent Adaptation Improves the Pick-and-Place Control of a Bioinspired Soft Grasper. Biomimetic and Biohybrid Systems, Living Machines 2024 14930:351-367.

Lyttle DN, Gill JP, Shaw KM, Thomas PJ, Chiel HJ (2017) Robustness, flexibility, and sensitivity in a multifunctional motor control model. Biological Cybernetics 111:25-47.

Mead C (2023) Neuromorphic Engineering: In Memory of Misha Mahowald. Neural Computation 35:343-383.

Mead CA (1989) Analog Vlsi and Neural Systems. Advanced Research in Vlsi : Proceedings of the Decennial Caltech Conference on Vlsi:1-1.

Minai AA (2024) Deep Intelligence: What AI Should Learn from Nature's Imagination. Cognitive Computation 16:2389–2404.

Minai AA (2025) Bounded Alignment: What (Not) To Expect From AGI Agents. In: 2025 IEEE/INNS International Joint Conference on Neural Networks. Rome, Italy.

Oakley BA (2012) Pathological Altruism. Oxford ; New York: Oxford University Press.

Oakley BA, Johnston M, Chen K, Jung E, Sejnowski T (2026) The Memory Paradox: Why Our Brains Need Knowledge in an Age of AI. In: The Artificial Intelligence Revolution: Challenges and Opportunities (Rangeley M, Fairfax N, eds). New York: Springer Nature.

Pugliese SM, Chou GM, Abe ETT, Turcu D, Lancaster JK, Tuthill JC, Brunton BW (2025) Connectome simulations identify a central pattern generator circuit for fly walking. bioRxiv:2025.2009.2012.675944.

Sejnowski TJ (2025) Dynamical Mechanisms for Coordinating Long-term Working Memory Based on the Precision of Spike-timing in Cortical Neurons. arXiv 2512.15891.

Sutton GP, Szczecinski NS, Quinn RD, Chiel HJ (2023) Phase shift between joint rotation and actuation reflects dominant forces and predicts muscle activation patterns. PNAS Nexus 2:pgad298.

Tam S, Hurwit I, Chiel HJ, Susswein AJ (2020) Multiple Local Synaptic Modifications at Specific Sensorimotor Connections after Learning Are Associated with Behavioral Adaptations That Are Components of a Global Response Change. Journal of Neuroscience 40:4363-4371.

Vaxenburg R, Siwanowicz I, Merel J, Robie AA, Morrow C, Novati G, Stefanidi Z, Both GJ, Card GM, Reiser MB, Botvinick MM, Branson KM, Tassa Y, Turaga SC (2025) Whole-body physics simulation of fruit fly locomotion. Nature 643.

Wan WE, Kubendran R, Schaefer C, Eryilmaz SB, Zhang WQ, Wu DB, Deiss S, Raina P, Qian H, Gao B, Joshi S, Wu HQ, Wong HSP, Cauwenberghs G (2022) A compute-in-memory chip based on resistive random-access memory. Nature 608:504-+.

Wang YY, Gill JP, Chiel HJ, Thomas PJ (2022) Variational and phase response analysis for limit cycles with hard boundaries, with applications to neuromechanical control problems. Biological Cybernetics 116:687-710.

Webster-Wood VA, Gill JP, Thomas PJ, Chiel HJ (2020) Control for multifunctionality: bioinspired control based on feeding in. Biological Cybernetics 114:557-588.